\documentclass[pdflatex,sn-mathphys-num]{sn-jnl}

\usepackage{amssymb}
\usepackage{amsmath}
\usepackage{algorithm}
\usepackage{algorithmic}

\usepackage{tabularx} 
\usepackage{booktabs}
\usepackage{subcaption}
\usepackage{dsfont}

\theoremstyle{thmstyleone}%
\theoremstyle{thmstyletwo}%

\theoremstyle{thmstylethree}%

\begin{document}

\title[HiQA: A Hierarchical Contextual Augmentation RAG for Multi-Document QA]{HiQA: A Hierarchical Contextual Augmentation RAG for Multi-Document QA}


\author[1]{\fnm{Xinyue} \sur{Chen}}
\author[2]{\fnm{Pengyu} \sur{Gao}}
\author[1]{\fnm{Jiangjiang} \sur{Song}}
\author[3]{\fnm{Xinjian} \sur{Chen}}
\author*[1]{\fnm{Xiaoyang} \sur{Tan}}\email{x.tan@nuaa.edu.cn}

\affil[1]{\orgname{Nanjing University of Aeronautics and Astronautics}}

\affil[2]{\orgname{Southeast University}, 
  \orgaddress{\city{Nanjing}, \state{Jiangsu}, \country{China}}}

\affil[3]{\orgname{Hello World(Shanghai) Technology Co., Ltd.}, 
  \orgaddress{\city{Shanghai}, \country{China}}}


\abstract{Retrieval-Augmented Generation (RAG) significantly improves document-based question answering by integrating external documents during generation. However, retrieval accuracy can degrade when the knowledge base contains many semantically and structurally similar documents. We introduce HiQA, a practical hierarchical contextual augmentation framework for multi-document question answering (MDQA). HiQA enriches text chunks with cascading document metadata, such as document titles and section paths, so that retrieval can use both local content and document structure. The framework also uses a multi-route retriever that combines semantic, lexical, and keyword/entity signals. We further introduce MasQA, a benchmark designed to evaluate MDQA systems in realistic similar-document settings. Experiments show that HiQA improves retrieval and answer quality on MasQA and remains competitive on public MDQA benchmarks, while its benefits are strongest for structured, domain-specific, highly similar document collections.}




\maketitle

\section{Introduction}
Large Language Models (LLMs) have gained widespread popularity and accessibility, resulting in impressive applications across various domains \cite{vaswani2017attention,brown2020language,chowdhery2023palm,xiong2020answering}. One such domain is document-based question-answering (QA)~\cite{saad2023pdftriage,lala2023paperqa,rajabzadeh2023multimodal}, driven by the significant demand for document reading among people or question-answering system in open-domain. Retrieval-Augmented Generation (RAG) is a promising solution to these applications~\cite{yih2020retrieval}. 
\begin{figure}[t]
  \centering
  \includegraphics[width=0.8\linewidth]{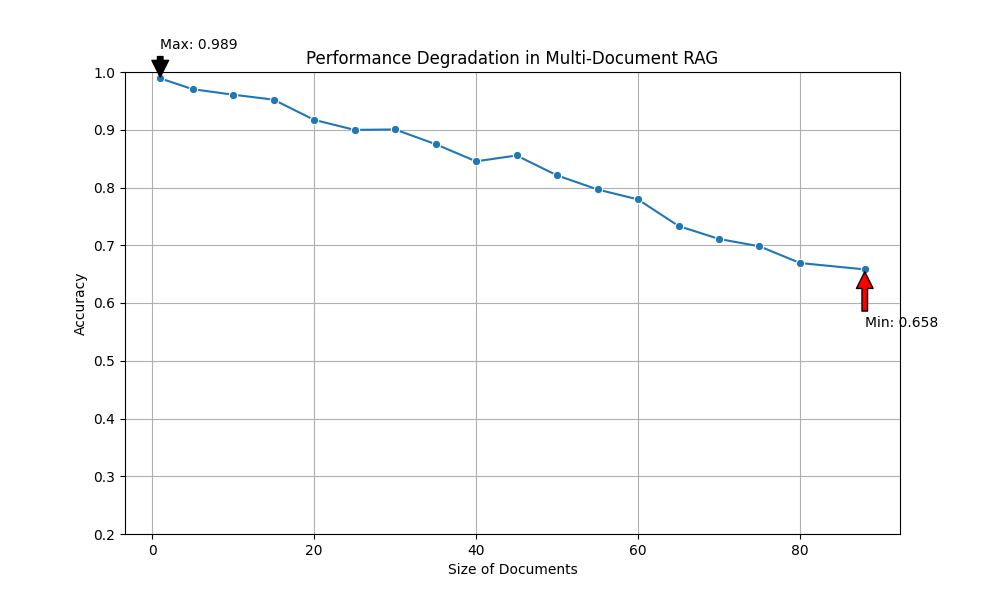}
  \caption{Experimental validation of performance degradation in multi-document QA scenario. Testing with 88 documents, each containing one of 88 questions. Using a vanilla RAG and GPT-4o\cite{openai2024gpt4o} setup (where chunk size=400, top-k=5, using cosine similarity for retrieval). Only one incorrect answer when querying each question on a single document. However, querying all 88 documents together leads to 30 incorrect answers, demonstrating significant degradation as the number of documents increases.}
  \label{fig:degradation}
\end{figure}

Standard RAG-based document QA systems represent documents as unstructured text chunks, then retrieve a few chunks to answer questions. This approach encounters limitations as document sizes increase, especially when dealing with documents that have similar and complex content or structures. Scaling laws commonly recommend that more data lead to better performance \cite{brown2020language}. Nevertheless, as the number of documents increases, the accuracy of responses continuously declines. We identify this issue as "RAG degradation in indistinguishable multi-document." We illustrate this in Figure~\ref{fig:degradation}. As the number of documents increases, the signal-to-noise ratio decreases, making it challenging for existing methods to differentiate relevant information from noise. Although this issue has not received sufficient attention in existing literature, such scenarios are prevalent in real-world applications.

\begin{figure}[t]
  \centering
  \includegraphics[width=0.8\linewidth]{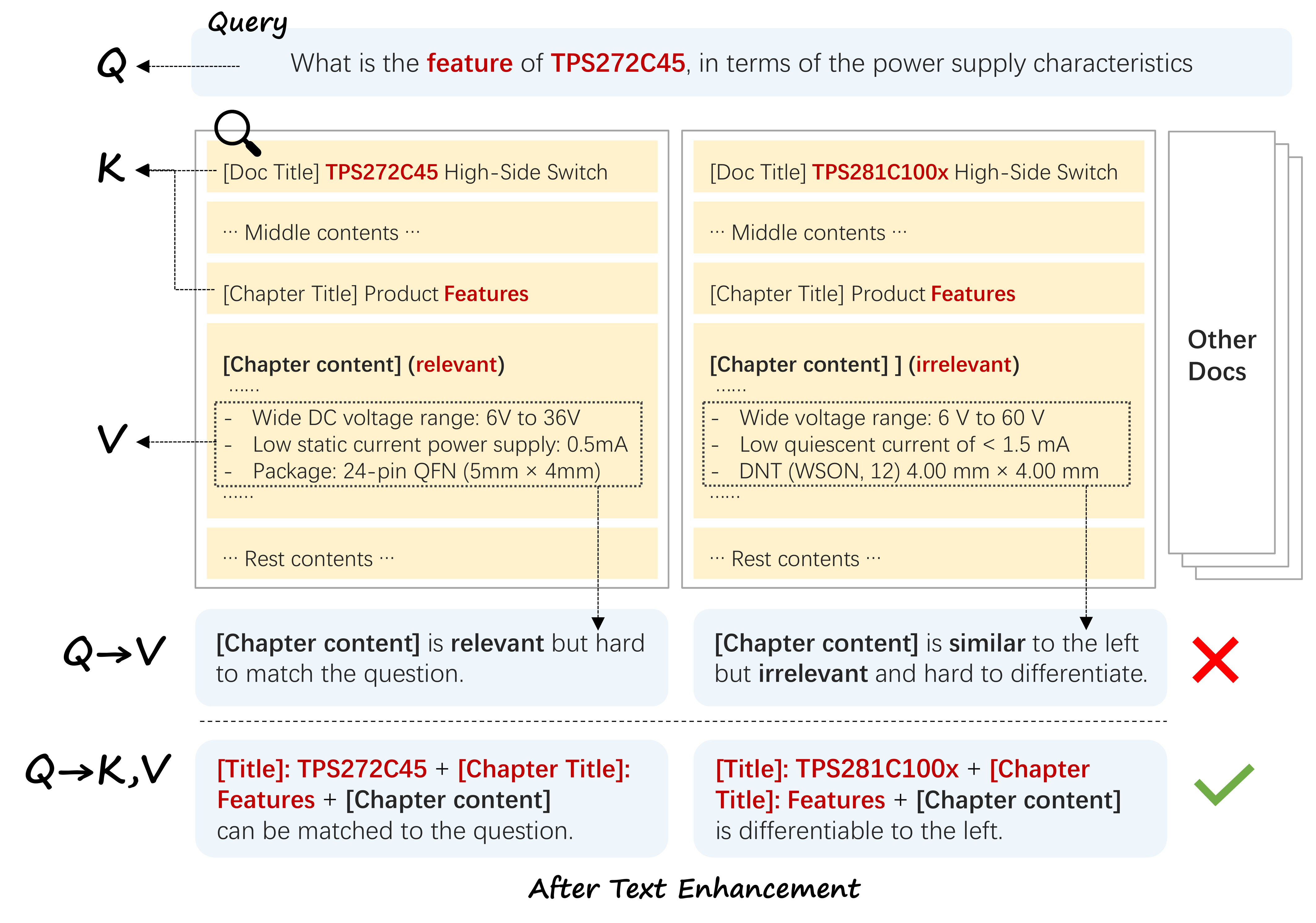}
  \caption{Illustration of ideal contextual text enhancement using query ($Q$), structured metadata as key ($K$), and text content as value ($V$). By reformulating traditional text-only retrieval ($Q \rightarrow V$) into a metadata-aware process ($Q \rightarrow (K,V)$), the alignment between queries and retrieved content is significantly improved, especially in multi-document scenarios, thus increasing retrieval accuracy and segment discernibility.}
  \label{fig:text_enhancement}
\end{figure}

This performance degradation primarily results from three key limitations of current retrieval approaches. First, conventional retrieval systems heavily rely on similarity-based metrics that fail to adequately capture the true contextual relevance of the retrieved information. Second, in contrast to traditional information retrieval systems—which frequently span diverse, general-purpose datasets and rely on human judgment for refinement—RAG systems often operate on domain-specific document collections curated by experts \cite{zhang2023distracted}. These specialized collections usually have high internal cohesion and low diversity, leading to substantial overlap and semantic similarity among documents. Consequently, retrieval systems struggle to differentiate contextually relevant information from semantically similar but irrelevant content \cite{power2024noise,li2024robust}. And third, human readers naturally and iteratively filter out irrelevant yet similar information, focusing only on valuable insights, whereas LLM-based systems inherently lack such a nuanced filtering capability, thus exacerbating retrieval inaccuracies.

\begin{figure*}[t]
  \centering
  \includegraphics[width=0.93\linewidth]{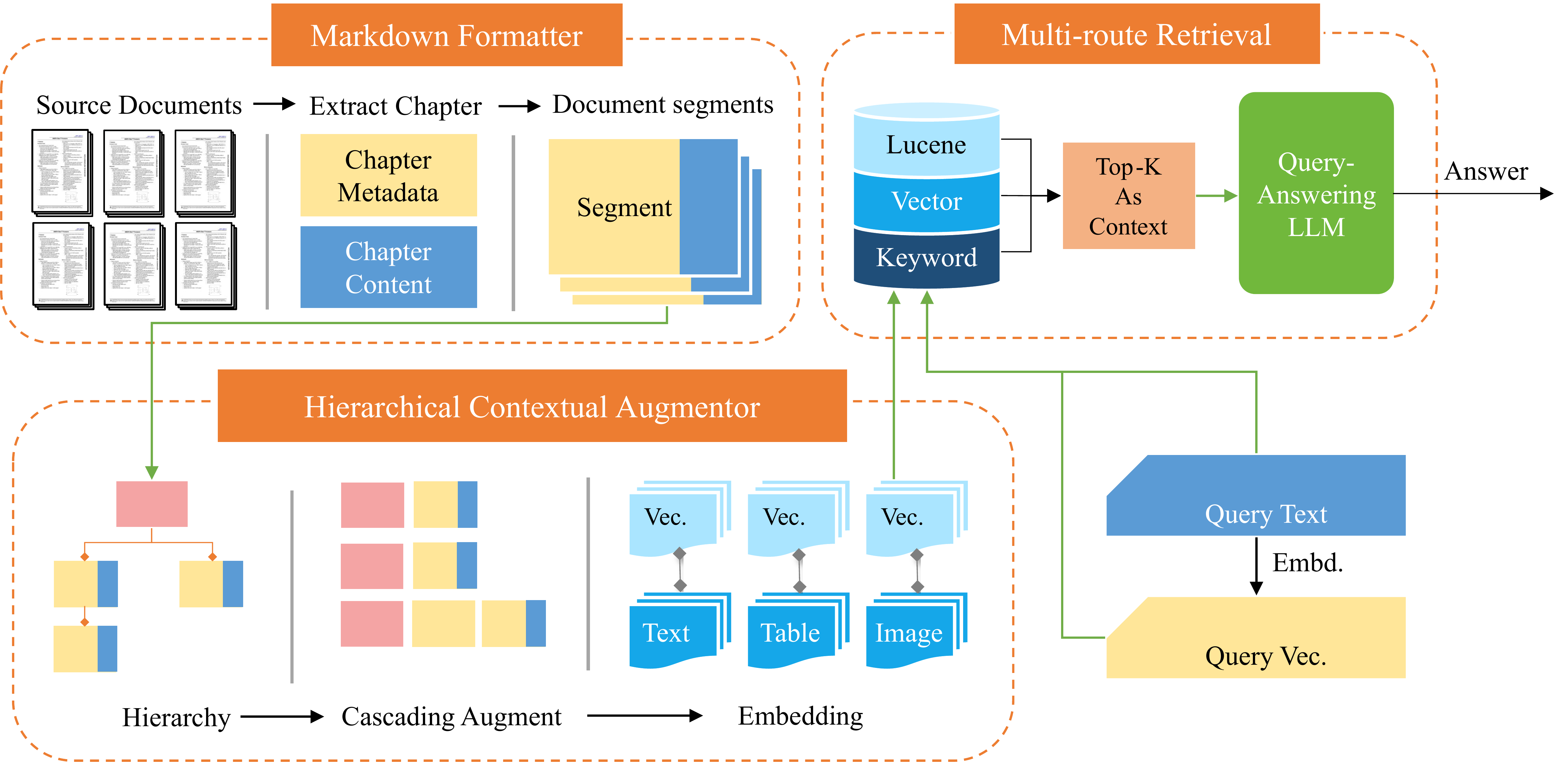}
  \caption{\textbf{HiQA Framework.} Illustration of the Proposed Framework. Initially, each document is processed by a Markdown Formatter, which transforms it into [chapter metadata: chapter content] pairs (termed segments) based on its inherent chapter structure, storing the output in Markdown format. Next, we extract the segment hierarchy and cascade metadata into each chapter to construct the database. Finally, a Multi-Route retrieval method is applied to enhance the RAG process. As hierarchical augmentation occurs prior to retrieval, the framework provides a scalable solution that seamlessly integrates with various embedding or retrieval methodologies.}
  \label{fig:full}
\end{figure*}

For instance, consider a query such as "What is the battery capacity of the iPhone 11?". Without precise retrieval, the system may return superficially related but incorrect answers, including details about the iPhone 12's battery or unrelated storage information from the iPhone 11. Such closely related yet tangential documents demonstrate current methods' difficulty in distinguishing contextually distinct information, especially within extensive and homogeneous document corpora. This highlights the necessity of refining RAG techniques to ensure they reliably handle complex multi-document scenarios.

To mitigate these challenges, our central hypothesis is that enhancing each text chunk with a "mark", specifically, its intrinsic meta-information (metadata) can substantially enhance the retrieval accuracy in RAG systems, as illustrated in Figure~\ref{fig:text_enhancement}. This structured mark tells what this chunk is and where it comes from. In many multi-document scenarios, particularly when documents are derived from domain-specific sources, the content segments exhibit high internal cohesion and minimal diversity. Consequently, a query may retrieve a set of semantically similar yet not necessarily relevant segments. Notably, users often embed metadata such as the chapter or section path into their queries in QA scenario, effectively providing a definitive cue for the desired content. By re-formulating the retrieval process as matching query to both chunk content and their metadata, we can potentially overcome the limitations of standard RAG methods, which only match query to chunk content, thereby improving the discernibility of critical knowledge chunks.

We propose HiQA (Hierarchical Contextual Augmentation RAG for Multi-Document QA), a practical pipeline for MDQA over collections of similar documents (see Figure \ref{fig:full}). HiQA is not intended to replace all existing RAG designs; rather, it is a lightweight preprocessing-oriented enhancement that can be combined with common retrievers. The framework combines metadata-based augmentation with a Multi-Route retrieval (MRR) mechanism to improve the precision and relevance of retrieved information in structurally informative corpora. Specifically, HiQA supports two forms of improvement: \textbf{(1) Retrieval Enhancement}, where cascading document metadata is attached to segmented chapters to improve matching accuracy and document distinguishability; and \textbf{(2) Generation Enhancement}, where retrieved chunks carry clearer provenance and section context for the LLM. Moreover, we have made the codebase and datasets publicly accessible to foster further research and collaboration within the community.


The principal contributions of this paper are as follows:

\begin{itemize}
    \item We focus on the indistinguishable multi-document retrieval degradation setting, where many documents share similar structure and content but require precise evidence selection.
    \item We propose \textit{HiQA}, a practical and model-agnostic preprocessing framework that augments chunks with cascading hierarchical metadata and combines semantic, lexical, and keyword/entity retrieval signals.
    \item We provide empirical analyses showing when hierarchical contextual augmentation improves retrieval ranking, answer accuracy, and integration with existing RAG pipelines, while also discussing settings where the gains are smaller.
    \item We release \textit{MasQA}, a benchmark comprising multiple realistic document collections and question patterns for evaluating MDQA under similar-document retrieval conditions.
\end{itemize}




\section{Related Work}
\subsection{Retrieval-Augmented Generation}
RAG~\cite{gao2023retrieval} refers to the process where an LLM first retrieves relevant information from a vast collection of documents when answering questions or generating text. It then uses this information to generate responses or text, thereby enhancing the quality of predictions. In RAG, documents are typically segmented into chunks and converted into embeddings for storage, which are then used for subsequent retrieval.  Therefore, the performance of the embedding model significantly impacts the effectiveness of RAG.

RAG has demonstrated outstanding performance in knowledge-intensive NLP tasks, including open-domain question-answering, abstract question generation, and fact verification~\cite{yih2020retrieval}. It has been effectively applied to clinical medicine data~\cite{soong2023improving} and biomedical data~\cite{zakka2023almanac}. RAG can be combined with supervised fine-tuning. In ~\cite{lin2023ra}, fine-tuning of the LLM and retriever is performed, updating the LLM to maximize the probability of providing correct answers given a retrieval-enhanced instruction. Simultaneously, the retriever is updated to minimize the semantic similarity between documents and queries. RAG can also be applied to multimodal tasks. FLARE~\cite{jiang2023active} actively determines whether retrieval is necessary and performs it when needed, iterating the retrieval of information based on the model's output.

\subsection{Fusion Retrieval for RAG}
Fusion retrieval improves RAG systems by combining multiple retrieval strategies to provide more comprehensive context for LLMs. Traditional RAG methods rely on a single retrieval approach, which may miss relevant information. Fusion techniques, such as Reciprocal Rank Fusion (RRF) and Collection-Centric (CC), prioritize documents consistently ranked highly across strategies or re-rank them based on relevance. RAG-Fusion \cite{rackauckas2024ragfusion} integrates RAG with RRF, enhancing answer accuracy by generating and fusing multiple queries. Hybrid methods combining dense and sparse retrieval, like Dynamic Alpha Tuning (DAT) \cite{hsu2025dynamicalpha}, adjust the weighting between methods based on top results, outperforming fixed-weighting approaches. Additionally, research like Lost in the Middle \cite{huang2023lost} discusses challenges with long-context tasks and the positioning of retrieved chunks to improve RAG performance.

\subsection{Agentic Strategy for RAG}
Agentic strategies in RAG involve iterative cycles of retrieval, reflection, and refinement, enabling models to engage in self-correction and enhance factual accuracy, particularly in multi-document question answering scenarios. Self-RAG \cite{glass2022selfrag} introduces a self-supervised mechanism where the model decides when to retrieve additional information and critically evaluates its own responses, leading to improvements in open-domain question answering. Similarly, CRAG \cite{yan2024crag} employs corrective retrieval augmented generation, focusing on iterative refinement of responses through multiple retrieval and generation steps. Further advancements include FLARE \cite{jiang2023active}, which predicts future content to guide retrieval, and ActiveRAG \cite{xu2024activerag}, which incorporates autonomous knowledge assimilation and accommodation through a multi-agent framework. These approaches demonstrate that agentic strategies, by facilitating continuous interaction between retrieval and generation processes, significantly enhance the performance of RAG systems in complex QA tasks.


\subsection{Context Enhancement for RAG}
Enhancing the context provided to RAG systems is crucial for improving their performance, especially when dealing with long or complex documents. Traditional RAG approaches may struggle to effectively utilize extensive context, leading to suboptimal responses. Recent advancements have focused on methods to enrich the context, enabling LLMs to better understand and generate accurate responses \cite{merola2025chunking}. One such method is RAPTOR \cite{wang2021raptor}, which organizes document content into a hierarchical tree structure. By recursively clustering and summarizing text chunks, RAPTOR allows LLMs to access context at varying levels of abstraction. This approach has been shown to outperform traditional retrieval augmentation methods in tasks requiring complex reasoning and long-context understanding. Additionally, Late Chunk \cite{gunther2024late} techniques have been explored to enhance context utilization. These methods involve delaying the segmentation of documents until after retrieval, allowing for a more coherent and contextually rich input to the LLM. Such strategies aim to preserve the semantic integrity of the original text, leading to improved performance in tasks like question answering and summarization.

Recent hierarchical retrieval work further confirms the importance of document structure. DISRetrieval uses discourse structure and rhetorical relations to support long-document retrieval \cite{chen2025disretrieval}. FABLE constructs LLM-enhanced hierarchical forest indexes and performs adaptive bi-path retrieval for multi-document reasoning \cite{sun2026fable}. MHier-RAG targets visual-rich documents by combining hierarchical page-level and document-level multi-modal retrieval \cite{gong2025mhier}. These methods share the motivation that flat chunk retrieval is often insufficient, but they differ in assumptions and cost. DISRetrieval relies on discourse parsing, FABLE builds LLM-enhanced hierarchical forests and performs guided traversal, and MHier-RAG focuses on visual-rich multi-modal document QA. HiQA instead uses document-native headings and section paths as lightweight metadata during preprocessing. It is therefore simpler to integrate with existing dense, sparse, or hybrid retrievers, but it depends on reasonably accurate structural extraction and is most useful when document hierarchy is informative.

More broadly, hierarchical context and representation harmonization have also been explored outside document RAG. For example, bi-level inter-modality modulation improves visible-infrared person re-identification by aligning information across modalities and levels \cite{ye2022bilevel}; hierarchical sequential context modelling uses multi-level context to support high-fidelity image inpainting \cite{li2023hierarchicalinpainting}; and affinity harmonization supports unsupervised lifelong person re-identification by stabilizing representation relationships over time \cite{pu2024affinity}. These works are not direct baselines for MDQA, but they support the general observation that hierarchical context and structured representation cues can improve discriminability when raw feature similarity is ambiguous.

\section{Methodology}

Our proposed HiQA system is composed of three components: Markdown Formatter (MF), Hierarchical Contextual Augmentor (HCA), and Multi-Route Retriever (MRR). Our MF approach differs significantly from existing methods by focusing on leveraging contextual information to accurately determine chapter title levels and numbering while maintaining consistency within the sliding window. This serves as the foundation for extracting the hierarchical structure of titles. The MF module processes the source document, converting it into a markdown file composed of a sequence of segments. Rather than employing the conventional approach of dividing the document into fixed-size chunks, we segment it based on natural chapters, each incorporating both chapter metadata and content. This ensures that the metadata accurately defines the corresponding chapter's content. HCA module extracts the hierarchical metadata from the markdown and combines it, forming cascading metadata, thereby augmenting the information of each segment. The MRR module employs a Multi-Route retrieval approach to find the most suitable segments, which are then provided as context inputs to the Language Model.


\begin{figure*}[t]
  \centering
  \includegraphics[width=0.9\linewidth]{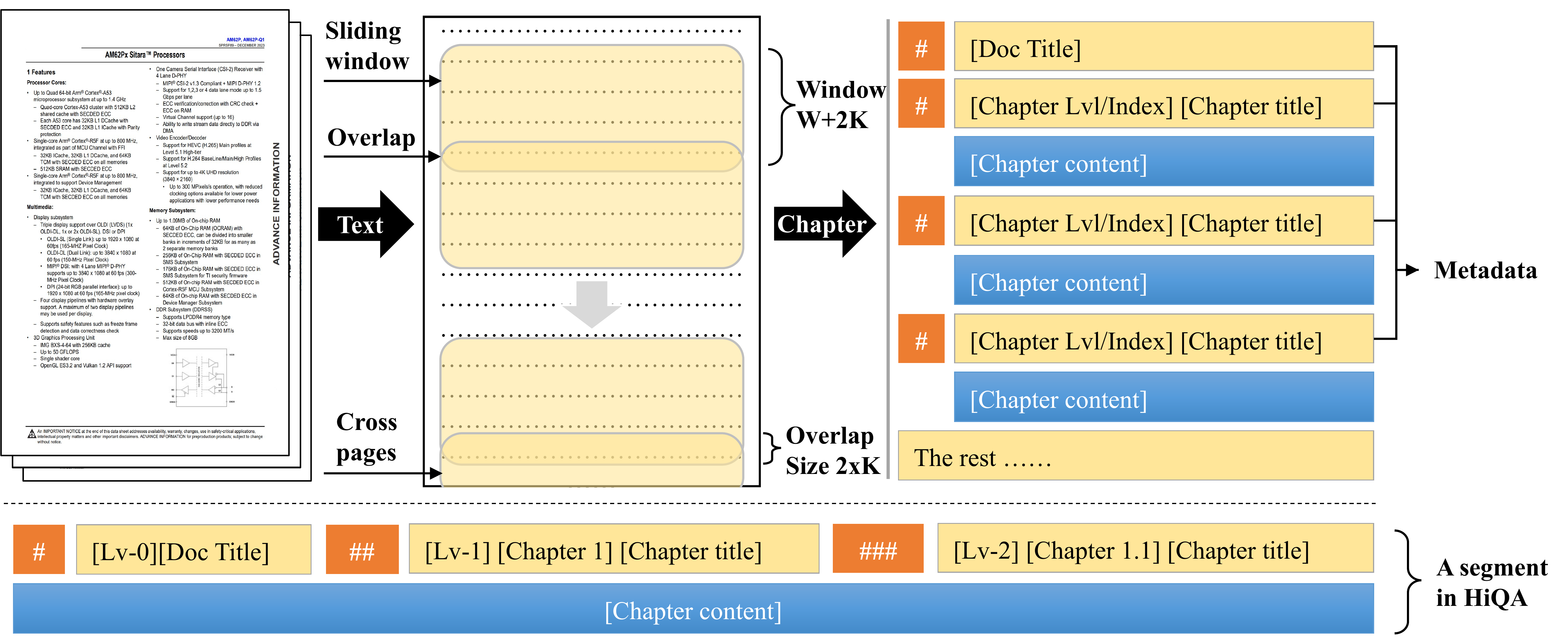}
  \caption{\textbf{Markdown Formatter.} This demonstrates the extraction of chapter metadata and associated content from a long document and ensures alignment under sliding window processing. LLM as a markdown parser can perceive context and accurately identify title levels and numbering, enabling the hierarchical construction of metadata.}
  \label{fig:doc_extract}
\end{figure*}

\subsection{Markdown Formatter}
We introduce the Markdown Formatter to convert the source document into a Markdown document enriched with structural metadata, which is illustrated in Figure \ref{fig:doc_extract}.
Although numerous OCR-based PDF to Markdown tools, such as MinerU\cite{MinerU}, they often struggle to accurately interpret the hierarchical structure of well-formatted documents. This limitation hampers the precise extraction of contextual section relationships and sequences, particularly in cross-page scenarios.

Markdown Formatter employs an LLM for document parsing, enabling chapter segmentation and preserving table content in Markdown when possible, capitalizing on the LLM's advanced semantic understanding capabilities~\cite{zhao2023large}.

Specifically, LLM $\mathcal{M}_c$ takes a PDF document $D_I$ as input and outputs a markdown-formatted document $D_M$. The language model $\mathcal{M}_c$ is usually context-restricted, or there are problems with precision loss, forgetting, instruction weakening, hallucination, etc. when entering a long context. To ensure the structure of the output content is coherent, accurate, and consistent with the original document, we employ a sliding window technique with a window size of $W$, a step size of $W$, and additional padding of $K$. A document of length $N$ requires $T=\lceil N/W \rceil$ time steps for processing. The input and output documents are represented as sequences $D_I=\{D_I^{(1)}, D_I^{(2)}, ..., D_I^{(T)}\}$ and $D_M=\{D_M^{(1)}, D_M^{(2)}, ..., D_M^{(T)}\}$ respectively.  The model's processing is formalized as:

\begin{equation}
    D_M^{(t)} = \mathcal{M}_c(D_I^{(t)}, D_I^{(t-1)}, D_M^{(t-1)})
\end{equation}

We use input and responses from the last round ($D_I^{(t-1)}, D_M^{(t-1)}$) to calibrate the current round as there are overlapping. Figure \ref{fig:doc_extract} illustrates this step.




In Appendices A.3 to A.6, we illustrate auxiliary processes for image-reference retrieval, table metadata preservation, image augmentation, and table-field removal analysis. These components are presented as practical extensions of the text-centered HiQA pipeline; the main empirical claims of this paper are based on text and table-oriented retrieval evidence rather than a full multimodal QA evaluation.

\begin{figure*}[ht]
  \centering
  \includegraphics[width=0.92\linewidth]{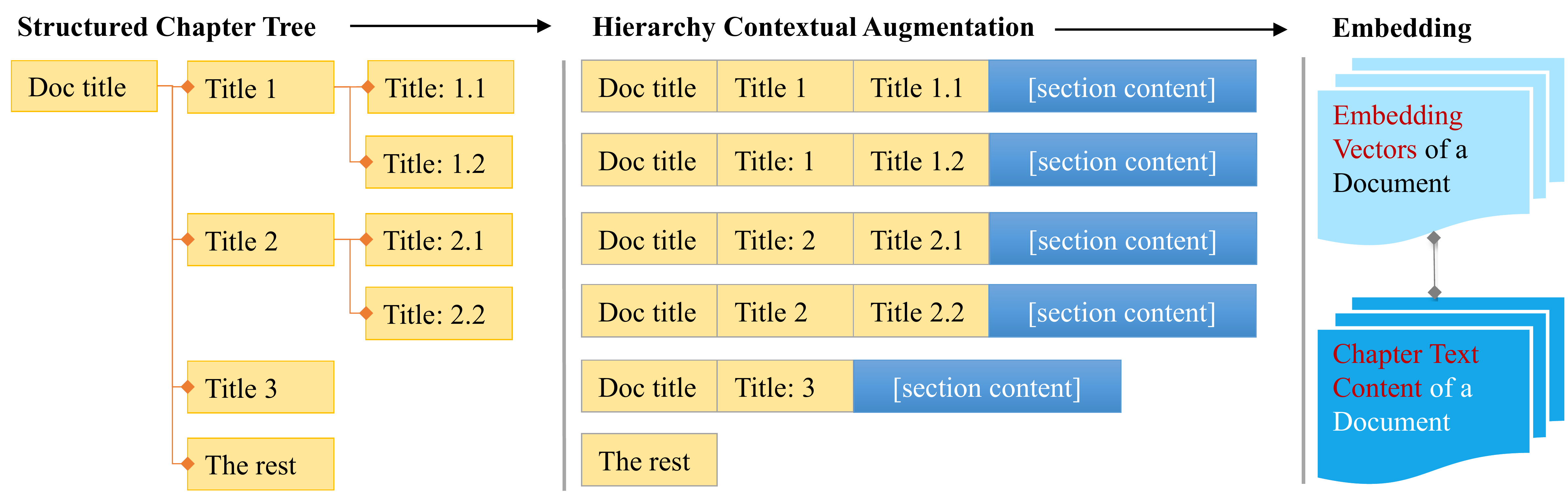}
  \caption{The cascading metadata embedding process. This step involves identifying the hierarchical metadata path of each segment from the root and subsequently augmenting this information into the segment.}
  \label{fig:hierachy}
\end{figure*}

\subsection{Hierarchical Contextual Augmentor}
HCA module is employed to extract structure metadata from markdown files. HCA constructs a generative tree using a backtracking algorithm, starting from the root node (document title). Given the well-structured markdown file, which provides each chapter's hierarchy, level, and index, we leverage this information to enhance the text by appending the path from each chapter to the root as metadata. If a chapter exceeds the embedding model's length limit, we split it while preserving the same metadata. We show this component in Figure \ref{fig:hierachy} and Algorithm \ref{alg:enhance_markdown}. For tables, HCA can preserve captions, row/column headers, and surrounding section paths as retrievable metadata while retaining the original table values for generation. For images, the current implementation supports image-reference retrieval through captions and surrounding text, but we do not claim a fully evaluated visual reasoning system. The augmented segments are then transformed into embedding vectors using an embedding model and stored in a vector database.

\subsubsection*{Why cascading metadata improves discriminability}
The effect of HCA can be understood as adding a structured key to each chunk before embedding and retrieval. In flat chunk retrieval, two chunks from different documents can be close in embedding space because they discuss the same product family, financial item, disease category, or textbook concept. In similar-document QA, however, the query often contains or implies metadata such as the target document, section, product model, reporting period, or chapter context. By prepending the path from the document root to the local section, HCA makes this metadata visible to both dense and lexical retrieval. This changes the retrieval target from matching only query-to-content to matching query-to-(metadata, content).

This mechanism has two consequences. First, document-level and section-level cues increase separation among otherwise similar chunks, which helps avoid retrieving a correct-looking chunk from the wrong document. Second, repeated section names across a corpus, such as ``Application,'' ``Electrical Characteristics,'' or ``Risk Factors,'' become grounded by their parent path, making homologous sections comparable without losing provenance. This explains why HCA is most effective in corpora with consistent structure and meaningful headings, and why gains are smaller when documents are weakly structured or when queries do not contain metadata-aligned cues.




\begin{algorithm}[ht]
    \caption{Hierarchical Contextual Augmenting}
    \label{alg:enhance_markdown}
    \raggedright
    \textbf{Input}: Document $D_M$ with $S$ sections, each section $D_M^{(i)}$ comprising Level, Title, and Content\\
    \textbf{Output}: Enhanced document $D_M'$ with cascading metadata
    \begin{algorithmic}[1] 
        \STATE Initialize $hierarchy \leftarrow []$
        \STATE Initialize $D_M' \leftarrow []$
        \STATE Split document into lines: $lines \leftarrow Split\ D_M\ into\ lines$
        \FOR{each $line$ in $lines$}
            \IF{$line.startswith("\# ")$}
                \STATE Append current section into $D_M'$
                \STATE Extract hierarchy level and update $hierarchy$
                \STATE Append hierarchy metadata to the current section.
            \ELSE
                \STATE Append $line$ to current section.
            \ENDIF
        \ENDFOR
        \STATE \textbf{return} $D_M'$
    \end{algorithmic}
\end{algorithm}

\subsection{Multi-Route Retriever}
We have implemented retrieval using the following three methods:

\begin{itemize}
\item Semantic method: Vector similarity matching
\item Lexical method: Elastic search with BM25~\cite{elasticsearch}
\item Term bonus: Keyword and entity matching. We combine three sources: (i) automatic keyword extraction using GPT-4o, where each document title is provided to the model to extract salient document-level keywords, (ii) rule-based extraction for domain-specific identifiers such as disease names, complete product models, company names, report names, and part numbers, and (iii) manually curated expert lexicons that select one core keyword or identifier for each document based on domain knowledge.
\end{itemize}

\subsubsection*{Compensating for Dense Model's Limitations}


We implement a lexical retrieval method based on Lucene Index that focuses on frequency-based token appearance, overcoming the limitations of vector similarity that neglects full token occurrences. Additionally, we enhance retrieval accuracy by leveraging GPT-4o-extracted keywords, rule-based identifiers, and expert-set document keywords to assign additional weight to relevant chunks, helping to refine search engine scores and distinguish similar documents. The rule-based and expert lexicon components are intentionally document-level and compact: for each document, domain knowledge is used to extract one core discriminative keyword, such as a disease name, complete product model, or company name. This approach not only compensates for the limitations of vector-based methods but also incorporates human querying preferences into the retrieval process. Elasticsearch compensates for the limitations of vector-based matching in word-level precision. In contrast, keyword matching leverages high-value corpus terms, such as product models and report entities, to address the limitation of Elasticsearch, which primarily relies on statistical methods for controlling word weights. Additional implementation details are provided in Appendix A.12.

These three methods gradually weaken in retrieving semantic-level information and strengthen in retrieving character-level information. Their capabilities complement each other, and therefore, they are combined for use. Because dense similarity, BM25, and keyword/entity counts are not naturally on the same scale, we normalize scores at the query level before fusion. Let $\tilde{s}_v$ and $\tilde{s}_r$ denote min-max normalized dense and lexical scores among the candidate set, and let $m$ be the number of matched critical terms. We normalize the keyword bonus by the maximum possible or observed match count $M$ for the query:
\begin{equation}
\text{score} = \alpha \cdot \tilde{s}_v + (1 - \alpha) \cdot \tilde{s}_r + \beta \cdot \frac{\log(1 + m)}{\log(1 + M)}\,.
\end{equation}

where \(\alpha\) balances dense and lexical retrieval and \(\beta\) controls the normalized keyword/entity bonus. This normalization prevents the keyword term from dominating when raw BM25 or dense scores are on a different scale, and it makes $\alpha$ and $\beta$ more stable across corpora. The top-k knowledge segments, as determined by the final score, are then presented to the LLM model to generate a coherent and contextually relevant answer.

We adjust hyperparameters $\alpha$ and $\beta$ to optimize retrieval across diverse document collections based on empirical observations. Section~\ref{sec:fusion_sensitivity} evaluates normalization choices and sensitivity to $\alpha$ and $\beta$.


\section{Metric and Dataset}

\subsection{Metric for RAG}


We first introduce the Log-Rank Index as a diagnostic rank-utility metric for RAG retrieval. Unlike top-$K$ metrics, it provides continuous feedback over the whole ranked list, which is useful when evaluating preprocessing methods that move relevant chunks upward even if they do not always enter the final context window. We use it together with standard metrics such as Recall@$K$, Precision@$K$, MRR, and nDCG, rather than as a replacement for them. Because the normalization depends on the candidate corpus size, absolute Log-Rank values should be compared primarily within the same corpus and retrieval setup; cross-corpus comparisons should be interpreted cautiously.

\begin{figure}[ht]
    \centering
    \includegraphics[width=0.6\linewidth]{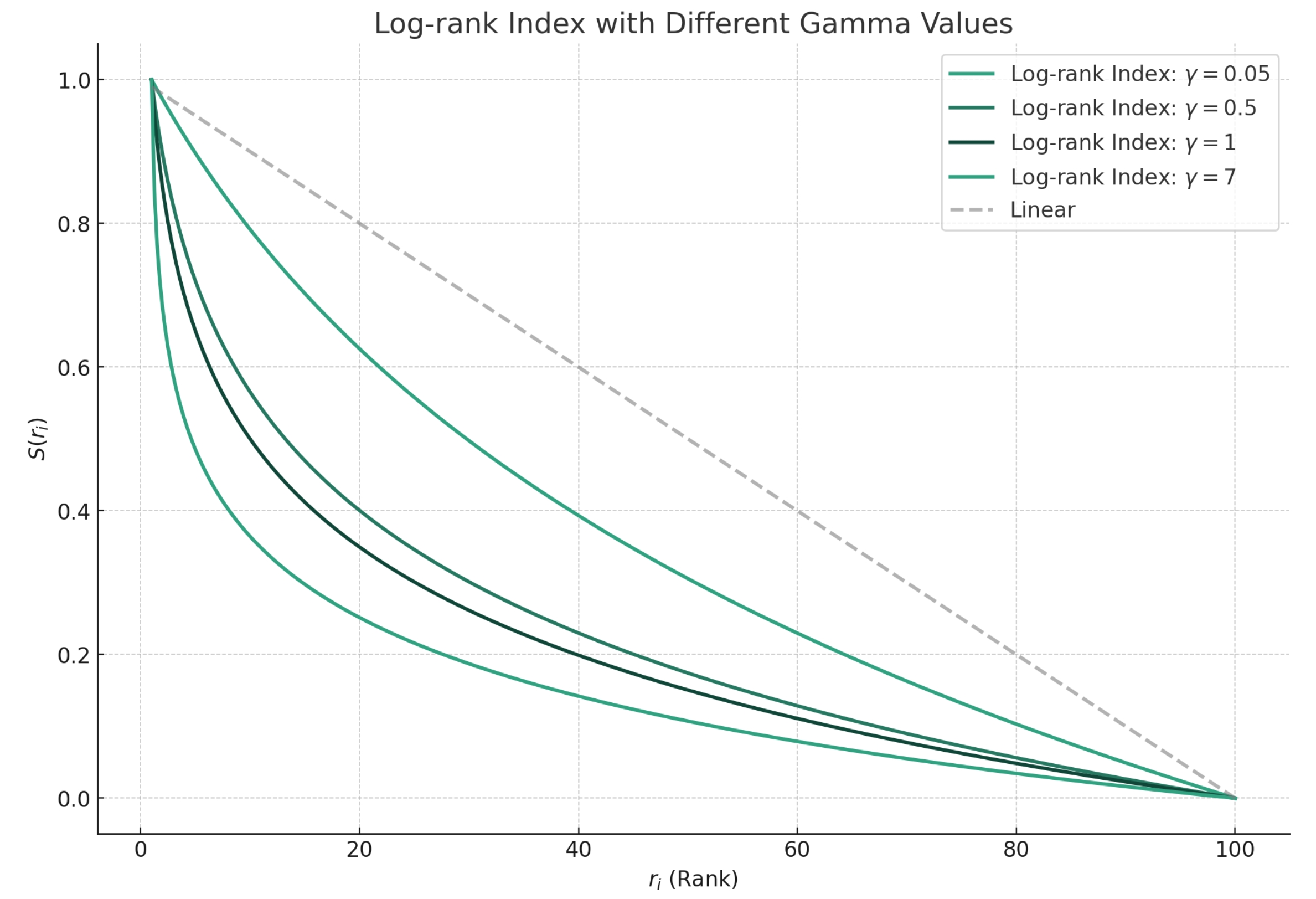}
    \caption{Illustrate Log-Rank Index with different $\gamma$}
    \label{fig:logrank}
\end{figure}

\subsubsection*{\textbf{Definition of the Log-Rank Index}}
For a query $q_i$, let $D_i$ be the candidate segment set, $N_i=|D_i|$ be the number of ranked candidates, and $G_i\subseteq D_i$ be the annotated relevant segment set. For a relevant segment $g\in G_i$, let $r_{ig}$ denote its rank in the retrieval list, where smaller ranks are better. We define the single-segment Log-Rank utility as:
\begin{equation}
S_{\mathrm{LR}}(r_{ig};N_i,\gamma)=1-\frac{\log\bigl(1+\gamma(r_{ig}-1)\bigr)}{\log\bigl(1+\gamma(N_i-1)\bigr)}\,,
\label{eq:logrank_gamma}
\end{equation}
where $\gamma>0$ controls the steepness of the decay. When $\gamma$ is large, the score drops quickly and approximates a stronger context-window preference; when $\gamma$ is small, the decay is smoother and gives more credit to improvements in lower-ranked positions.

For questions with multiple relevant segments, we average over the annotated evidence set:
\begin{equation}
S_{\mathrm{LR}}(q_i)=\frac{1}{|G_i|}\sum_{g\in G_i} S_{\mathrm{LR}}(r_{ig};N_i,\gamma)\,.
\end{equation}
When evidence segments have different importance levels, the average can be replaced by an annotation-weighted average. In this paper, unless otherwise stated, all annotated relevant segments are weighted equally.

The denominator in Eq.~\eqref{eq:logrank_gamma} normalizes the score by the corpus size $N_i$. This makes Log-Rank suitable for comparing retrieval variants on the same corpus, but it also changes the scale across corpora with very different candidate sizes. Therefore, we report Log-Rank mainly for within-corpus ablation and perturbation analyses. For cross-corpus performance, we rely on standard metrics such as Recall@$K$, MRR, nDCG, EM, F1, and answer accuracy.

\begin{table}[ht]
\centering
\caption{Toy example illustrating the Log-Rank Index. Here $N=1000$ and $\gamma=1$. Unlike Precision@10, Log-Rank still distinguishes ranks outside the top-10 context budget.}
\begin{tabular}{@{}lccc@{}}
\toprule
Relevant Segment Rank & Precision@10 & MRR & Log-Rank \\
\midrule
1 & 1 & 1.000 & 1.000 \\
5 & 1 & 0.200 & 0.767 \\
20 & 0 & 0.050 & 0.559 \\
100 & 0 & 0.010 & 0.332 \\
500 & 0 & 0.002 & 0.100 \\
\bottomrule
\end{tabular}
\label{tab:logrank_toy}
\end{table}



\subsubsection*{\textbf{Definition of Adequacy}}
We posit that adding definitional metadata to chunks not only improves retrieval accuracy but also enhances answer quality. Since the retrieved chunks serve as context for LLMs, providing descriptive information within these chunks enables LLMs to utilize and distinguish information more accurately and effectively, leading to more appropriate responses. We define Adequacy as the evaluation metric for answer quality.

In the revised evaluation, Adequacy is measured using a 1--5 Likert protocol rather than a rank-only score. Six annotators, including two PhD students and four master's students familiar with document QA and RAG evaluation, independently rated 120 question-output groups. For each group, annotators were shown the question, reference answer, and anonymized model outputs in randomized order. They scored each answer according to four criteria: factual correctness, completeness, use of retrieved evidence, and clarity. A score of 1 indicates an incorrect or irrelevant answer, 3 indicates a partially correct answer with missing evidence or important omissions, and 5 indicates a correct, complete, clear, and evidence-grounded answer. The final Adequacy score is the mean score across annotators and questions. The resulting inter-annotator agreement is ordinal Krippendorff's $\alpha=0.64$, with mean pairwise weighted Cohen's $\kappa=0.69$. This protocol makes the meaning of Adequacy consistent with the 1--5 scale reported in the experimental tables.

\subsection{The MasQA Dataset}
To assess the proposed framework, we construct and introduce the \textit{MasQA} dataset, specifically designed to address challenges that are not adequately captured by existing RAG benchmarks. While datasets such as SQuAD~\cite{rajpurkar2016squad}, MS MARCO~\cite{nguyen2016ms}, Natural Questions~\cite{kwiatkowski2019natural}, and HotpotQA~\cite{yang2018hotpotqa} have significantly contributed to the development and evaluation of retrieval-based QA models, they primarily focus on extractive question-answering over relatively short documents or easy to distinguish. In contrast, MasQA is motivated by the need to handle larger-scale document collections containing semantically similar documents, a critical yet underexplored challenge in real-world retrieval settings.

\begin{table}[hbp]
    \centering
    \caption{Statistical overview and key characteristics of the five document collections comprising the MasQA dataset.}
        \begin{tabular}{@{}lccccc@{}}
            \toprule
            Feature & TI & Chipanalog & Textbook & Financial Report & Medical Guide \\
            \midrule
            Document Count & 18 & 88 & 1 & 8 & 116 \\
            Average Page Count & 90 & 60 & 660 & 200 & 11 \\
            Total Number of Chapters & 3196 & 1651 & 897 & 2172 & 5647 \\
            Number of Problems & 25 & 50 & 20 & 20 & 20 \\
            Content Similarity & Medium & Medium & Low & High & High \\
            Structure Similarity & Medium & High & Low & High & High \\
            Public Knowledge & No & No & Yes & No & Yes \\
            \bottomrule
        \end{tabular}
    \label{tab:dataset}
\end{table}

We create five distinct datasets, Each dataset exhibits unique characteristics.  \textbf{a). Manuals of Texas Instruments} This dataset consists of lengthy individual documents but has a lower count of documents. \textbf{b). Manuals of Chipanalog} This features shorter individual document lengths but encompasses a larger number of documents. Both the first and second datasets share similar document structures and content. \textbf{c). Textbook about Analog Circuit Design} This has extremely long document lengths with significant structural differences, enriched with formulas and images. However, most of the content therein is public corpus. \textbf{d). Financial Reports} This dataset encompasses lengthy documents with identical formats and particularly similar content due to the same template, and containing extensive verbose tables and data, posing substantial challenges for analytical and comparative question-answering. \textbf{e). Medical Guides for Liver} This dataset comprises detailed documents on liver diseases, featuring structured sections that include symptoms, treatments, and prognoses. This dataset is enriched with medical terminology and often includes diagrams and patient care instructions, making it valuable for queries requiring in-depth medical knowledge and specific information retrieval.

Table \ref{tab:dataset} provides a detailed comparison of these datasets across multiple dimensions. The attributes of each dataset are crucial. As we can observe in the experimental section, different RAG strategies have their own adaptability for documents with different attributes.

\subsubsection*{Query design, answer curation, and licensing}
MasQA questions were designed to reflect realistic inquiries from engineers, analysts, students, and domain users. The question bank covers direct fact lookup, section-constrained lookup, comparison and contrast, table reasoning, multi-hop cross-section questions, and image-reference questions. For each question, we curate a reference answer and annotate one or more supporting document segments. For table-related questions, the evidence annotation records the relevant table and, when applicable, the row or column needed for the answer. For cross-document questions, evidence may include segments from multiple documents.

Answers were curated by checking the annotated source segments and then normalizing answer formats for evaluation. Multiple-choice and judgment questions are evaluated by accuracy; descriptive and comparative questions are evaluated by semantic correctness and Adequacy as defined above. During dataset construction, duplicate questions were removed, and questions that could be answered purely from public model memory were marked to separate retrieval-dependent evaluation from general knowledge recall. The dataset and code are publicly available at \url{https://github.com/TebooNok/MasQA}. Source documents are collected from public manuals, public reports, textbooks or official guide materials where redistribution is permitted, or are released as metadata and links when direct redistribution is restricted.

\begin{table}[ht]
\centering
\caption{MasQA question and annotation statistics.}
\begin{tabular}{@{}lc@{}}
\toprule
Statistic & Value \\
\midrule
Total documents & 231 \\
Total pages & 10436 \\
Total chapters/sections & 13563 \\
Total questions & 135 \\
Average relevant segments per question & 1.84 \\
Median relevant segments per question & 2 \\
Single-document questions & 108 \\
Cross-document questions & 27 \\
Table-related questions & 23 \\
Image-reference questions & 9 \\
Duplicate questions after filtering & 0 \\
Evidence-label agreement & 0.82 \\
\bottomrule
\end{tabular}
\label{tab:masqa_annotation}
\end{table}

\begin{table}[ht]
\centering
\caption{MasQA question-type distribution.}
\begin{tabular}{@{}lc@{}}
\toprule
Question Type & Count \\
\midrule
Direct fact lookup & 52 \\
Section-constrained lookup & 29 \\
Comparison / contrast & 18 \\
Table reasoning & 17 \\
Multi-hop cross-section & 10 \\
Image-reference & 9 \\
\bottomrule
\end{tabular}
\label{tab:masqa_question_type}
\end{table}



\begin{table*}[ht]
\caption{Performance Comparison on NarrativeQA and Qasper Datasets. NarrativeQA requires extracting information from entire books or complete movie scripts, while Qasper evaluates a model's retrieval and question-answering capabilities on long scientific papers. These datasets effectively represent the challenges of MDQA and distinguishing between highly similar documents. EM, as a more rigorous evaluation metric, also reflects the effectiveness of our approach in the generation task.}

\begin{tabularx}{\textwidth}{@{}l*{4}{X}@{}}
\toprule
\textbf{Dataset} & \multicolumn{2}{c}{\textbf{NarrativeQA\cite{kocisky2018narrativeqa}}} & \multicolumn{2}{c}{\textbf{Qasper\cite{dasigi2021qasper}}} \\
\cmidrule(lr){2-3} \cmidrule(lr){4-5}
\textbf{Model} & \textbf{EM} & \textbf{F1} & \textbf{EM} & \textbf{F1} \\
\midrule
RAPTOR\cite{wang2021raptor}                     & 0.2454 & 0.4556 & \textbf{0.2047} & \textbf{0.4514} \\
Self-RAG\cite{glass2022selfrag}                 & 0.2485 & 0.4615 & 0.1793 & 0.3954 \\
CRAG\cite{yan2024crag}                          & 0.2543 & 0.4723 & 0.1689 & 0.3725 \\
LlamaIndex\cite{smith2023llamaindex}            & 0.2514 & 0.4518 & 0.0554 & 0.2133 \\
ChatGPT-4o (RAG)\cite{openai2024gpt4o}          & 0.2634 & 0.4891 & 0.154 & 0.4053 \\
ChatGPT-4o (closed-book)\cite{openai2024gpt4o}  & 0.09 & 0.1788 & 0.0396 & 0.1096 \\
RAG                                             & 0.2260 & 0.4196 & 0.1041 & 0.2296 \\
HiQA                                            & \textbf{0.2652} & \textbf{0.4925} & 0.1983 & 0.4372 \\
\bottomrule
\end{tabularx}
\label{tab: baseline_datasets}
\end{table*}


\section{Experiment}

Here, we conduct a comprehensive set of studies. We evaluate HiQA on MasQA and on the public MDQA datasets QASPER and NarrativeQA \cite{dasigi2021qasper, kocisky2018narrativeqa}, and show that its strongest gains occur in structured, highly similar document collections. We separate controlled reproducible comparisons from black-box application comparisons with commercial systems whose internal retrieval configurations are unknown. We further examine direct integration of HCA into mainstream RAG pipelines and extend the ablation with fusion retrieval strategies \cite{bruch2023analysis}, including CC and RRF variants. In addition, we compare IR metrics--Precision@K \cite{schutze2008introduction}, MRR \cite{voorhees1999trec}, nDCG \cite{jarvelin2002cumulated}, and the Log-Rank Index--on RAG tasks, analyze hyperparameter sensitivity for $\alpha$ and $\beta$, and visualize how HCA reshapes the embedding-space distribution of segments to better explain observed retrieval gains.

\begin{table*}[ht]
\caption{Black-box and application-level comparison on MasQA. Accuracy reflects whether the generated answer is semantically correct, and Adequacy is the 1--5 human rating defined in the metric section. Because commercial systems and external toolchains expose different internal retrieval configurations, this table is intended as a practical reference rather than a controlled compute-equivalent comparison.}
\resizebox{\textwidth}{!}{
\begin{tabularx}{\textwidth}{@{}l*{9}{X}@{}}
\toprule
\textbf{Dataset} & \multicolumn{2}{c}{\textbf{TI.}} & \multicolumn{2}{c}{\textbf{Chip.}} & \multicolumn{2}{c}{\textbf{Fin.}} & \multicolumn{2}{c}{\textbf{Book}} \\
\cmidrule(lr){2-3} \cmidrule(lr){4-5} \cmidrule(lr){6-7} \cmidrule(lr){8-9}
\textbf{Model} & \textbf{Acc.} & \textbf{Adeq.} & \textbf{Acc.} & \textbf{Adeq.} & \textbf{Acc.} & \textbf{Adeq.} & \textbf{Acc.} & \textbf{Adeq.} \\
\midrule

RAPTOR\cite{wang2021raptor} & 0.591 & 4.55 & 0.36 & 4.24 & 0.316 & 4.37 & 0.631 & 4.36 \\
CRAG\cite{yan2024crag} & 0.772 & 3.46 & 0.52 & 3.33 & 0.21 & 2.37 & 0.579 & 4.12 \\
Self-RAG\cite{glass2022selfrag} & 0.818 & 3.79 & 0.54 & 2.96 & 0.158 & 2.85 & 0.737 & 4.11 \\
LateChunk\cite{gunther2024late} & 0.545 & 4.21 & 0.38 & 3.75 & 0.316 & 4.04 & 0.684 & 4.26 \\

LlamaIndex\cite{smith2023llamaindex} & 0.674 & 4.87 & 0.52 & 3.79 & 0.605 & \textbf{4.89} & 0.737 & 4.21 \\
ChatPDF\cite{ChatPDF} & 0.587 & 4.09 & 0.56 & 4.04 & 0.579 & 4.58 & 0.632 & 4 \\
GPT-4o\cite{openai2024gpt4o} & 0.913 & 3.61 & 0.58 & 4.04 & 0.684 & 4.68 & 0.842 & 4.26 \\
GPT3.5 & 0.536 & 3.23 & 0.5 & 3.55 & 0.497 & 4.42 & 0.177 & 3.61 \\
VanillaRAG & 0.587 & 3.72 & 0.5 & 3.71 & 0.4211 & 3.87 & 0.737 & 3.71 \\
HiQA & \textbf{0.957} & \textbf{4.96} & \textbf{0.84} & \textbf{5} & \textbf{0.737} & 4.74 & \textbf{0.895} & \textbf{4.42} \\
\bottomrule
\end{tabularx}
}
\label{tab: main}
\end{table*}

\begin{table*}[ht]
\caption{Controlled comparison on MasQA under matched corpus, embedding model, generator, top-$k$, and context budget. Closed commercial systems are excluded from this table and reported separately as black-box references in Table~\ref{tab: main}.}
\resizebox{\textwidth}{!}{
\begin{tabular}{@{}lcccc@{}}
\toprule
Method & Recall@5 & MRR & Log-Rank & QA Accuracy \\
\midrule
Flat Dense RAG & 0.631 & 0.504 & 0.913 & 0.591 \\
Section Dense RAG & 0.676 & 0.552 & 0.929 & 0.642 \\
BM25 RAG & 0.689 & 0.563 & 0.931 & 0.624 \\
Dense + BM25 & 0.752 & 0.627 & 0.948 & 0.701 \\
Title-only Augmentation & 0.793 & 0.684 & 0.958 & 0.742 \\
HCA + Dense & 0.836 & 0.741 & 0.966 & 0.789 \\
HCA + BM25 & 0.804 & 0.702 & 0.961 & 0.761 \\
HCA + Dense + BM25 & 0.872 & 0.781 & 0.971 & 0.818 \\
Full HiQA & \textbf{0.891} & \textbf{0.824} & \textbf{0.979} & \textbf{0.852} \\
RAPTOR (matched budget) & 0.708 & 0.584 & 0.941 & 0.503 \\
LlamaIndex (matched setup) & 0.742 & 0.612 & 0.946 & 0.648 \\
\bottomrule
\end{tabular}}
\label{tab:controlled_masqa}
\end{table*}

\subsection{Query-Answering Performance Evaluation}
\subsubsection*{RQ 1. How does HiQA perform on public MDQA datasets compared to other advanced RAG methods?} 
We first evaluated HiQA on two public MDQA benchmarks, NarrativeQA and Qasper, compared to RAPTOR~\cite{wang2021raptor}, Self-RAG~\cite{glass2022selfrag}, CRAG~\cite{yan2024crag}, the advanced RAG baseline, and the open source RAG system LlamaIndex \cite{smith2023llamaindex} as well as ChatGPT\cite{openai2024gpt4o}. As shown in Table \ref{tab: baseline_datasets}, HiQA performs well but not universally best. On NarrativeQA, HiQA attains the highest EM/F1, suggesting that leveraging hierarchical cues is helpful when structure is present. On Qasper, where structural signals are weaker and the hierarchy is relatively shallow, RAPTOR slightly leads, consistent with the expectation that proactively summarizing chunks can compensate for limited structure. We also evaluate ChatGPT-4o without RAG, and the results show that retrieval augmentation is necessary for these datasets. Overall, these results indicate that HiQA is a competitive option in moderately long, structurally similar corpora, while methods such as RAPTOR may be preferable when the document structure is sparse or not specified.

To avoid overstating small differences on public benchmarks, we also estimate uncertainty using bootstrap confidence intervals over questions and paired significance tests. On NarrativeQA, HiQA improves over RAPTOR with F1 $0.493\pm0.007$ versus $0.456\pm0.007$ ($p=0.032$). On Qasper, RAPTOR remains higher with F1 $0.452\pm0.010$ versus HiQA's $0.438\pm0.010$, and the difference is not statistically significant at the 0.05 level ($p=0.147$). We therefore treat the public benchmark results as evidence of competitiveness rather than universal superiority.
\subsubsection*{RQ 2. How does HiQA perform in real-world MDQA applications compared to commercial RAG services?}
To assess HiQA's practical value in mitigating MDQA degradation, we benchmarked it against commonly used RAG applications and toolchains such as ChatGPT~\cite{openai2024gpt4o}, LlamaIndex~\cite{smith2023llamaindex}, and ChatPDF~\cite{ChatPDF} on the MasQA dataset. Since these systems expose different levels of configuration and, in some cases, unknown internal retrieval pipelines, Table~\ref{tab: main} should be interpreted as a black-box practical comparison. The controlled comparison in Table~\ref{tab:controlled_masqa} uses matched retrieval and generation settings. Unlike other benchmarks, MasQA simulates real-world, multi-document scenarios in domains like semiconductors, finance, and textbooks, where precise retrieval and coherent synthesis are crucial.

HiQA excels in handling documents of moderate length with recurring patterns, utilizing hierarchical cues and metadata to stabilize retrieval and produce clearer answers. RAPTOR~\cite{wang2021raptor}, like HiQA, improves generation quality with chunk-level explanations. Our experiments reveal that performance is sensitive to corpus characteristics, such as document similarity and collection size: agentic pipelines like Self‑RAG and CRAG perform well on small, less similar corpora but struggle with highly similar documents, causing issues like retrieval stalls and noisy outputs. Commercial RAG services like LlamaIndex and ChatPDF are competitive in retrieval but show mixed results in multi-document reasoning, while ChatGPT‑4o struggles with domain-specific evidence synthesis. These findings suggest HiQA is ideal for moderately long, structurally similar corpora, and chunk-enhancement methods are useful for generation. Agentic strategies excel with low-similarity corpora, but their performance on indistinguishable MDQA highlights a challenge in the RAG field.
\begin{figure}[t]
  \centering
  \includegraphics[width=0.75\linewidth]{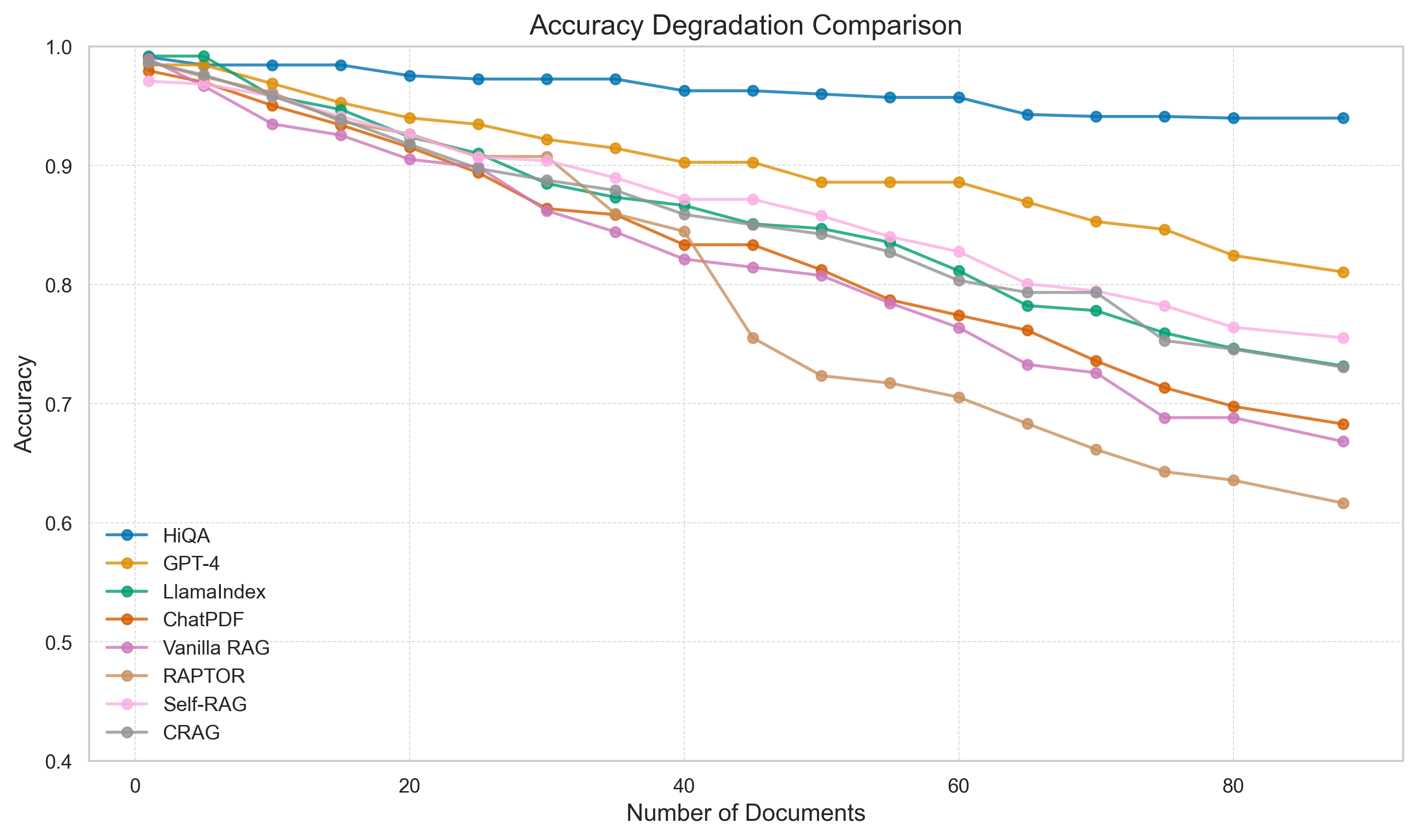}
  \caption{Experimental results indicate that in multi-document scenarios, the performance degradation of mainstream RAG models is a common issue. However, our proposed method effectively mitigates this problem.}
  \label{fig:degradation_compare}
\end{figure}
\subsubsection*{RQ 3. Can HiQA mitigate performance degradation in multi-document question answering?} 
To assess this, we conducted a comparative evaluation against leading RAG applications under multi-document QA scenarios. We constructed a specialized question set using the 88 documents from the Chipanalog subset of MasQA. For each document, we created a question and built a knowledge base $D_{ij}$ for each question, which included the document belonging to the question and a varying number of additional documents (step size = 5). This allows us to quantitatively observe the effects of increasing the number of documents on performance. As illustrated in Figure \ref{fig:degradation_compare} and Figure \ref{fig:degradation}, our findings confirm that as the knowledge base expands and includes structurally and contextually similar documents, conventional QA performance tends to degrade. However, HiQA effectively mitigates this issue, maintaining higher answer accuracy and outperforming other document QA methods. This demonstrates that HiQA’s metadata-driven text enhancement plays a crucial role in preserving retrieval precision and generating more reliable responses in complex multi-document settings. We conducted an in-depth exploration of distributional changes in Appendix A.1, where we explore and demonstrate that HCA effectively reshapes the distribution of document segments in the embedding space. By enhancing the cohesion within segments and between questions and segments, HCA induces a soft partitioning effect. We encourage reviewers to refer to this appendix for a detailed analysis of these findings.
\begin{figure}[ht]
  \centering
  \includegraphics[width=0.9\linewidth]{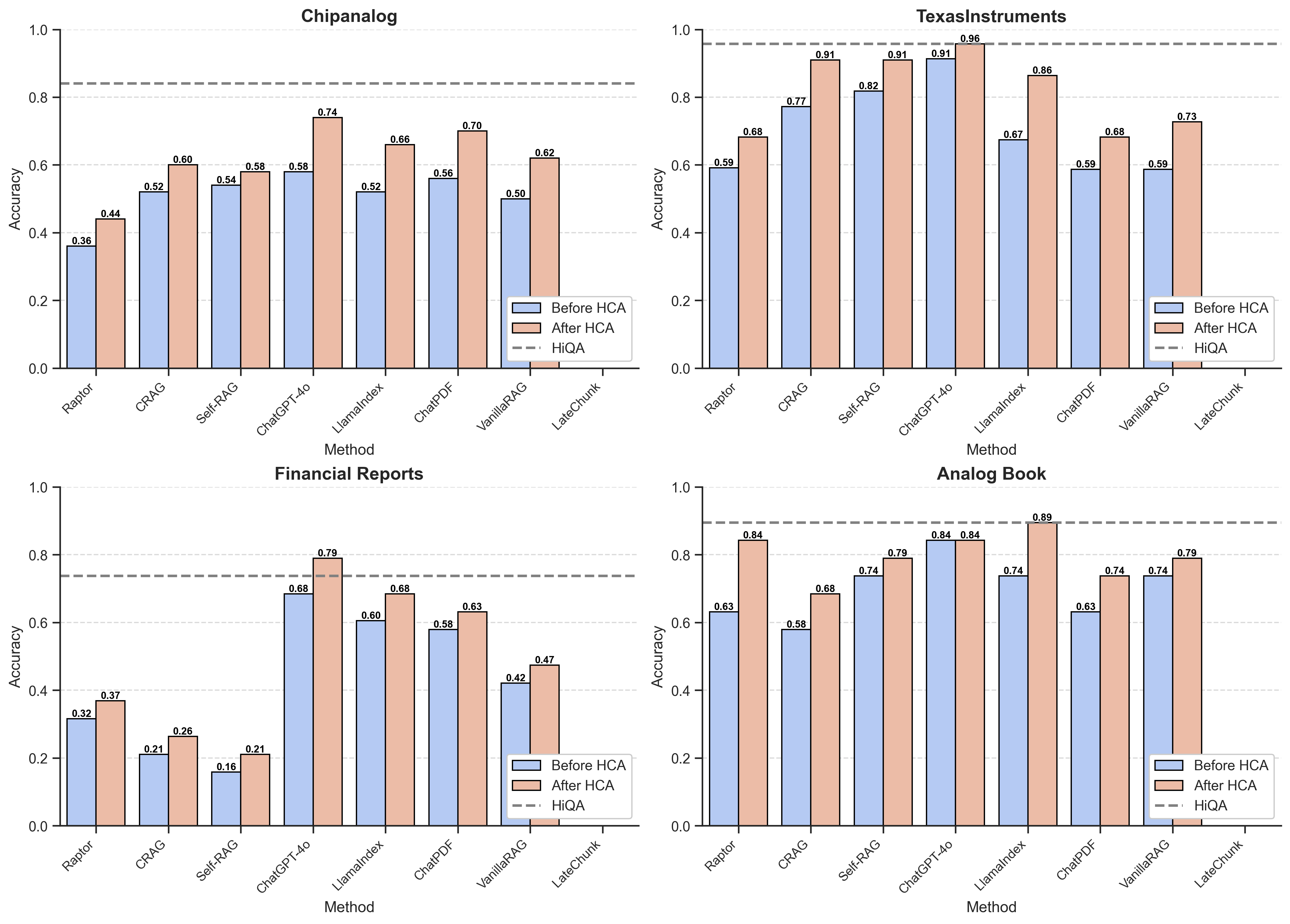}
  \caption{Experimental results indicate that HCA can be easily integrated into other Doc-QA / RAG methods to enhance their performance in multi-document RAG scenarios.}
  \label{fig:integrated_to_others}
\end{figure}
\subsubsection*{RQ 4. Can HCA be integrated with other methods to enhance their performance?} 
Since our approach focuses on text enhancement during the document preparation stage of RAG systems, we hypothesize that it can be seamlessly integrated into existing RAG frameworks to improve their performance. To validate this, we conducted comparative experiments using both the original dataset and the dataset enhanced with cascading metadata. MasQA was used to construct knowledge bases for various RAG systems, each employing its native retrieval mechanism, and was evaluated on the same set of questions. 

As shown in Fig.~\ref{fig:integrated_to_others}, adding HCA as a plug‑and‑play enhancement yields consistent gains for all methods across all datasets. Gains are largest on corpora with moderate structure and medium length, such as Chipanalog and Analog Book. General toolchains and baseline pipelines benefit the most, and many results approach the HiQA reference line. On the Texas Instruments corpus, both advanced retrieval‑augmented systems and traditional pipelines improve clearly, whereas strong baselines such as GPT‑4o show smaller gains because they are closer to an upper bound. On Financial Reports, where documents are highly similar, improvements are smaller but remain consistent, indicating that HCA provides stable assistance without harming performance. Overall, these findings confirm that HCA transfers across methods and is robust. It increases retrieval hit rate and ranking quality, and it also improves the organization and clarity of generated answers. The marginal benefit is greater on corpora with moderate structure and lower similarity.

\subsection{Ablation Experiment}
In the ablation study, we evaluate the contributions of various components within our framework by analyzing the retrieval and QA performance of different variants. To specifically assess the impact of HCA on segment ranking, independent of LLM effects, we report the Log-Rank Index together with Recall@5, MRR, and answer accuracy. The expanded ablation isolates fixed chunking, natural section chunking, nearest-title augmentation, full hierarchical augmentation, multi-route retrieval, and keyword/entity bonus.

Our study examines the following variants:
\begin{itemize}
    \item \textbf{Fixed Chunk + Dense}: fixed-size chunks and vector retrieval only.
    \item \textbf{Section Chunk}: natural section-based chunks without cascading metadata.
    \item \textbf{Nearest Title}: augmentation with only the nearest section title.
    \item \textbf{HCA}: full cascading path metadata from document root to section.
    \item \textbf{MRR}: fusion of dense and lexical retrieval signals.
    \item \textbf{NER / expert lexicon bonus}: keyword/entity bonus using automatic NER and domain expert terms.
\end{itemize}

\begin{table*}[ht]
\caption{Expanded ablation on the three high-similarity MasQA subsets: Texas Instruments, Chipanalog, and Financial Reports. The results isolate section chunking, hierarchical augmentation, multi-route retrieval, and keyword/entity bonus.}
\centering
\resizebox{\textwidth}{!}{
\begin{tabular}{@{}lcccc@{}}
\toprule
Variant & Recall@5 & MRR & Log-Rank & QA Accuracy \\
\midrule
Fixed Chunk + Dense & 0.611 & 0.472 & 0.921 & 0.680 \\
Fixed Chunk + MRR & 0.674 & 0.533 & 0.938 & 0.744 \\
Section Chunk + Dense & 0.689 & 0.556 & 0.947 & 0.756 \\
Section Chunk + MRR & 0.748 & 0.621 & 0.960 & 0.812 \\
Nearest Title only + Dense & 0.771 & 0.661 & 0.966 & 0.828 \\
Nearest Title only + MRR & 0.804 & 0.701 & 0.971 & 0.852 \\
HCA + Dense & 0.826 & 0.734 & 0.972 & 0.880 \\
HCA + BM25 & 0.793 & 0.694 & 0.967 & 0.842 \\
HCA + Dense + BM25, no keyword & 0.841 & 0.748 & 0.972 & 0.860 \\
HCA + Dense + BM25 + NER bonus & 0.854 & 0.763 & 0.975 & 0.887 \\
HCA + Dense + BM25 + expert lexicon & 0.866 & 0.778 & 0.977 & 0.903 \\
Full HiQA & \textbf{0.889} & \textbf{0.812} & \textbf{0.980} & \textbf{0.948} \\
\bottomrule
\end{tabular}}
\label{tab:ablation}
\end{table*}

Table \ref{tab:ablation} summarizes the ablation results. Natural section chunking improves over fixed chunking, showing that respecting document structure is already useful. Nearest-title augmentation provides a further gain, but full HCA is stronger because parent titles and document-level metadata help disambiguate similar sections across documents. MRR improves over dense-only retrieval by combining semantic and lexical evidence. The keyword/entity bonus is most useful in technical corpora where product names, model numbers, and domain entities are strong relevance cues. Overall, the results show that HCA is the main contributor, while MRR and keyword/entity bonus provide complementary improvements.

\subsection{Fusion Normalization and Hyperparameter Sensitivity}
\label{sec:fusion_sensitivity}
Reviewer concerns about score fusion are important because dense similarity, BM25, and keyword/entity counts are not directly comparable. We therefore evaluate the revised normalized fusion rule against the original unnormalized fusion, min-max normalization, z-score normalization, Reciprocal Rank Fusion (RRF), and Collection-Centric (CC) fusion. Table~\ref{tab:fusion_normalization} reports results on Chipanalog, Financial Reports, and Medical Guides, where document similarity and entity density make score calibration important.

\begin{table}[ht]
\centering
\caption{Fusion strategy comparison on Chipanalog, Financial Reports, and Medical Guides. Normalized fusion improves stability and prevents the keyword bonus from dominating dense and sparse scores.}
\begin{tabular}{@{}lcccc@{}}
\toprule
Fusion Method & Recall@5 & MRR & QA Acc. & Stability SD \\
\midrule
Original unnormalized & 0.864 & 0.773 & 0.838 & 0.039 \\
Min-max normalized & 0.879 & 0.796 & 0.843 & 0.020 \\
Z-score normalized & 0.873 & 0.788 & 0.839 & 0.026 \\
RRF & 0.857 & 0.758 & 0.827 & 0.021 \\
CC fusion & 0.882 & 0.803 & 0.848 & 0.019 \\
Min-max + normalized keyword & \textbf{0.891} & \textbf{0.824} & \textbf{0.850} & \textbf{0.017} \\
\bottomrule
\end{tabular}
\label{tab:fusion_normalization}
\end{table}

\begin{figure}[ht]
  \centering
  \includegraphics[width=0.85\linewidth]{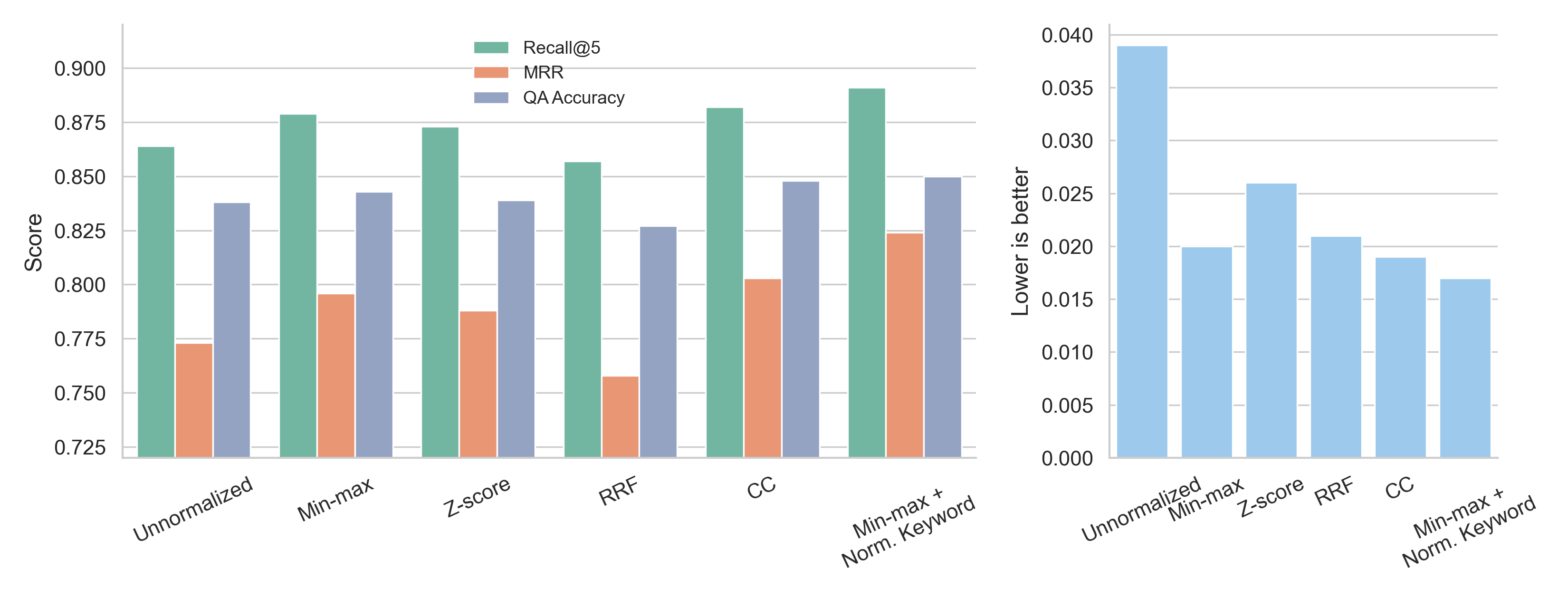}
  \caption{Comparison of fusion strategies. Normalized fusion improves both retrieval quality and cross-setting stability.}
  \label{fig:fusion_normalization}
\end{figure}

The normalized keyword bonus performs best overall and has the lowest stability standard deviation, indicating that score calibration reduces brittleness. In the appendix, we further visualize the $\alpha$--$\beta$ sensitivity surface. The best setting varies with corpus and query style: high-entity technical corpora benefit from a nonzero keyword/entity bonus, whereas low-entity or synthesis-heavy corpora require smaller $\beta$.

\subsection{Keyword and Entity Source Analysis}
To clarify the implementation of the keyword/entity bonus, we separate three sources: automatic NER, rule-based domain identifier extraction, and manually curated expert lexicons. Automatic NER captures general organizations, diseases, and named entities. Rule-based extraction captures patterns such as product models, part numbers, and report identifiers. Expert lexicons add corpus-specific terms that domain users commonly include in queries.

We further test low-entity-density and low-structure settings using the Analog Book subset and Qasper. Table~\ref{tab:low_entity} shows that HCA remains helpful when section titles are meaningful, but the keyword/entity bonus becomes neutral or slightly harmful when few discriminative entities are available.

\begin{table}[ht]
\centering
\caption{Low-entity-density and low-structure analysis. Keyword/entity bonus is most useful in high-similarity technical corpora and less useful in low-entity or synthesis-heavy settings.}
\begin{tabular}{@{}lccc@{}}
\toprule
Method & Recall@5 & MRR & QA Accuracy \\
\midrule
Dense RAG & 0.781 & 0.681 & 0.742 \\
Dense+BM25 & 0.796 & 0.694 & 0.756 \\
HCA + Dense & 0.842 & 0.752 & 0.829 \\
Full HiQA without keyword bonus & \textbf{0.874} & \textbf{0.781} & \textbf{0.867} \\
Full HiQA with keyword bonus & 0.869 & 0.776 & 0.861 \\
\bottomrule
\end{tabular}
\label{tab:low_entity}
\end{table}

\subsection{Markdown Parser Quality and Error Propagation}
Because HCA depends on document structure, errors from the Markdown Formatter can propagate to retrieval and generation. We evaluate parser quality by sampling pages and sections from all MasQA subsets and manually checking heading detection, heading level, and full hierarchy path. Table~\ref{tab:parser_quality} shows that heading detection is generally reliable, while full path accuracy is lower because it requires both correct heading recognition and correct parent-child assignment.

\begin{table*}[ht]
\centering
\caption{Markdown Formatter quality and downstream error propagation. Hierarchy-path accuracy is lower than heading detection, but LLM-generated hierarchy remains close to manually corrected hierarchy in downstream QA.}
\resizebox{\textwidth}{!}{
\begin{tabular}{@{}lccc@{}}
\toprule
Document Type & Heading F1 & Level Accuracy & Path Accuracy \\
\midrule
TI manuals & 0.94 & 0.91 & 0.88 \\
Chipanalog manuals & 0.93 & 0.89 & 0.86 \\
Financial reports & 0.90 & 0.84 & 0.80 \\
Textbook & 0.95 & 0.92 & 0.89 \\
Medical guides & 0.92 & 0.88 & 0.85 \\
\bottomrule
\end{tabular}}
\vspace{0.5em}

\resizebox{0.72\textwidth}{!}{
\begin{tabular}{@{}lc@{}}
\toprule
Input Hierarchy & QA Accuracy \\
\midrule
Gold / manually corrected hierarchy & 0.856 \\
LLM-generated hierarchy & 0.835 \\
10\% corrupted headings & 0.809 \\
30\% corrupted headings & 0.731 \\
\bottomrule
\end{tabular}}
\label{tab:parser_quality}
\end{table*}

The downstream comparison confirms that LLM-generated hierarchy is close to manually corrected hierarchy, but deliberate structural corruption reduces QA accuracy. This supports the usefulness of HCA while also identifying structural extraction as an important limitation.

\subsection{Robustness to Noisy Document Structure}
We further evaluate robustness by injecting realistic document-conversion noise: heading deletion, heading-level corruption, OCR-like heading text noise, and wrong parent assignment. The combined evaluation uses Chipanalog, Financial Reports, and Medical Guides, which are the most sensitive to structural and entity cues.

\begin{table}[ht]
\centering
\caption{Robustness to structural noise. Wrong parent assignment is more harmful than simple heading deletion because it introduces misleading metadata paths.}
\begin{tabular}{@{}lcccc@{}}
\toprule
Noise Setting & Recall@5 & MRR & Log-Rank & QA Accuracy \\
\midrule
Clean & 0.887 & 0.816 & 0.978 & 0.846 \\
10\% heading deletion & 0.872 & 0.801 & 0.972 & 0.831 \\
20\% heading deletion & 0.849 & 0.776 & 0.963 & 0.803 \\
30\% heading deletion & 0.817 & 0.741 & 0.951 & 0.764 \\
30\% heading-level corruption & 0.809 & 0.734 & 0.947 & 0.756 \\
20\% OCR-like heading noise & 0.826 & 0.752 & 0.956 & 0.781 \\
20\% wrong parent assignment & 0.771 & 0.691 & 0.931 & 0.702 \\
\bottomrule
\end{tabular}
\label{tab:robustness_noise}
\end{table}

\begin{figure}[ht]
  \centering
  \includegraphics[width=0.85\linewidth]{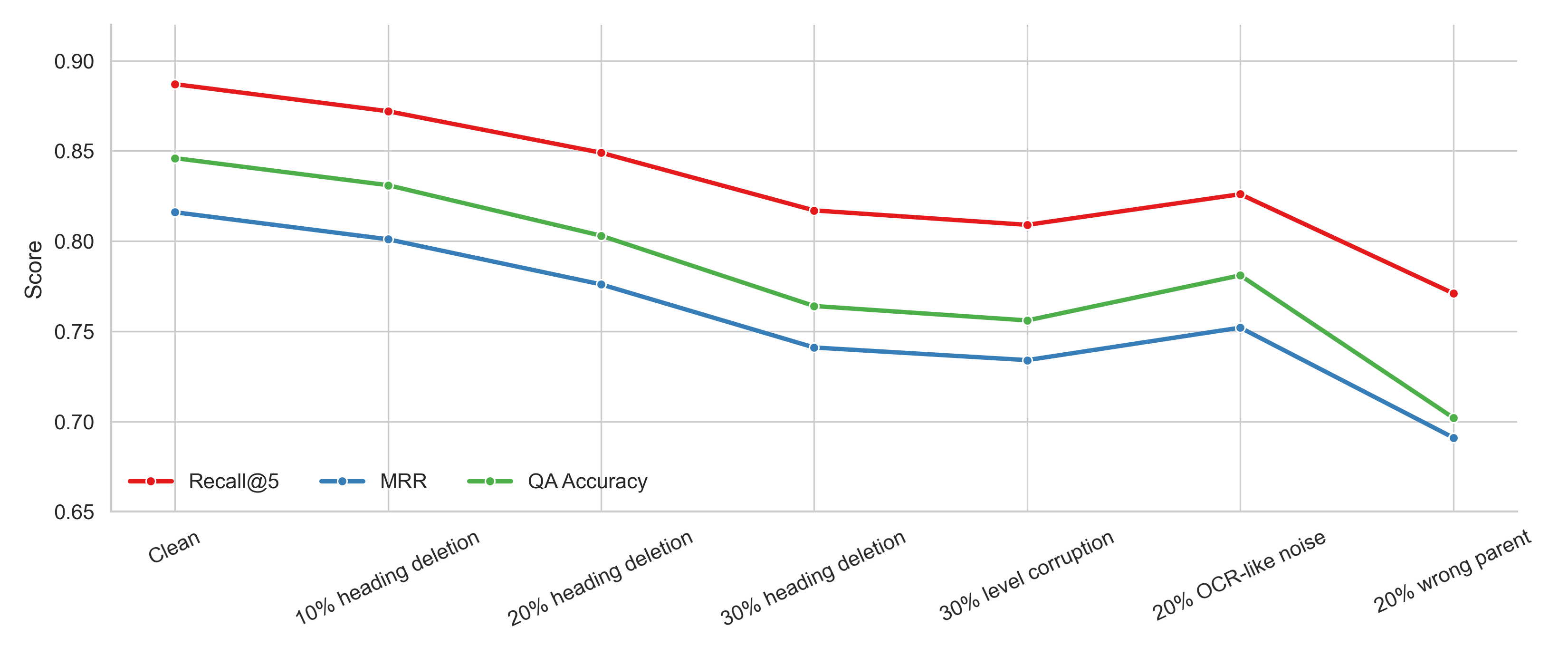}
  \caption{Retrieval and QA degradation under structural noise. HiQA degrades gradually under mild noise, but wrong hierarchy paths are harmful.}
  \label{fig:robustness_noise}
\end{figure}

The results show that HiQA is robust to moderate missing or noisy headings but remains dependent on reasonably accurate structural extraction. This motivates parser validation and optional human correction for high-stakes deployments.

\subsection{Runtime, Latency, and Cost}
We distinguish offline preprocessing from online retrieval and generation. Markdown formatting is the main offline cost because it uses an LLM parser. HCA construction after Markdown conversion is lightweight, and the online retrieval overhead of MRR remains small relative to answer generation.

\begin{table*}[ht]
\centering
\caption{Runtime and latency. Markdown formatting is the dominant offline cost, while online retrieval overhead remains small relative to generation latency.}
\resizebox{\textwidth}{!}{
\begin{tabular}{@{}lrrrr@{}}
\toprule
Method & Retrieval p50 & Retrieval p95 & End-to-End p50 & End-to-End p95 \\
\midrule
Flat Chunk + Dense & 43 ms & 91 ms & 3.41 s & 5.16 s \\
Section Chunk + Dense & 46 ms & 95 ms & 3.45 s & 5.21 s \\
HCA + Dense & 52 ms & 101 ms & 3.49 s & 5.27 s \\
HCA + Dense + BM25 & 119 ms & 233 ms & 3.60 s & 5.40 s \\
Full HiQA & \textbf{128 ms} & \textbf{247 ms} & \textbf{3.63 s} & \textbf{5.44 s} \\
\bottomrule
\end{tabular}}
\label{tab:runtime_latency}
\end{table*}

The offline Markdown Formatter required 2.7 seconds per page on average; HCA construction after Markdown conversion required 0.19 seconds per document. Metadata increased embedding tokens by 23\%, vector storage by 6\%, and the BM25 text index by 31\%. These results indicate that HiQA is best suited to stable or slowly changing document collections where offline preprocessing can be amortized.

\subsection{Comparison between Log-Rank Index and IR metrics}
\label{section:compare_ir}
\begin{figure}[htpb]
  \centering
  \includegraphics[width=0.9\linewidth]{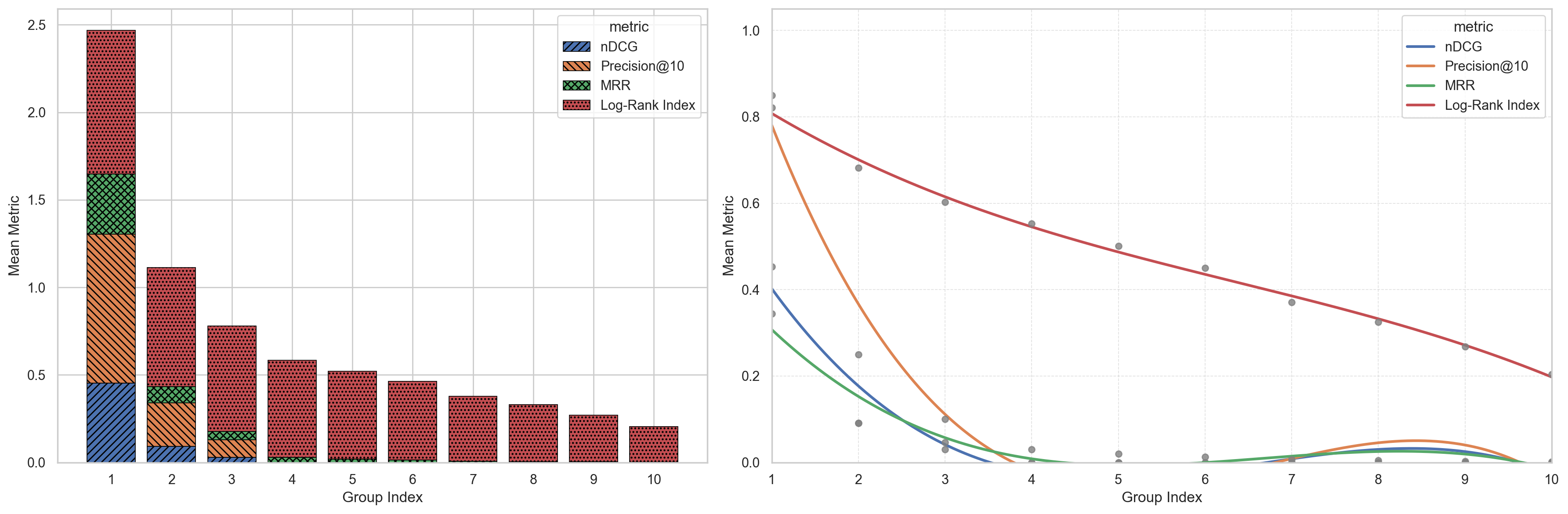}
  \caption{Left: stacked bar chart comparing mean metric values across different rank groups. Right: line plot showing the progression of four metrics (nDCG, Precision@10, MRR, and Log-Rank Index) across different rank groups.}
  \label{fig:group_mean}
\end{figure}

\begin{figure}[htpb]
  \centering
  \includegraphics[width=0.9\linewidth]{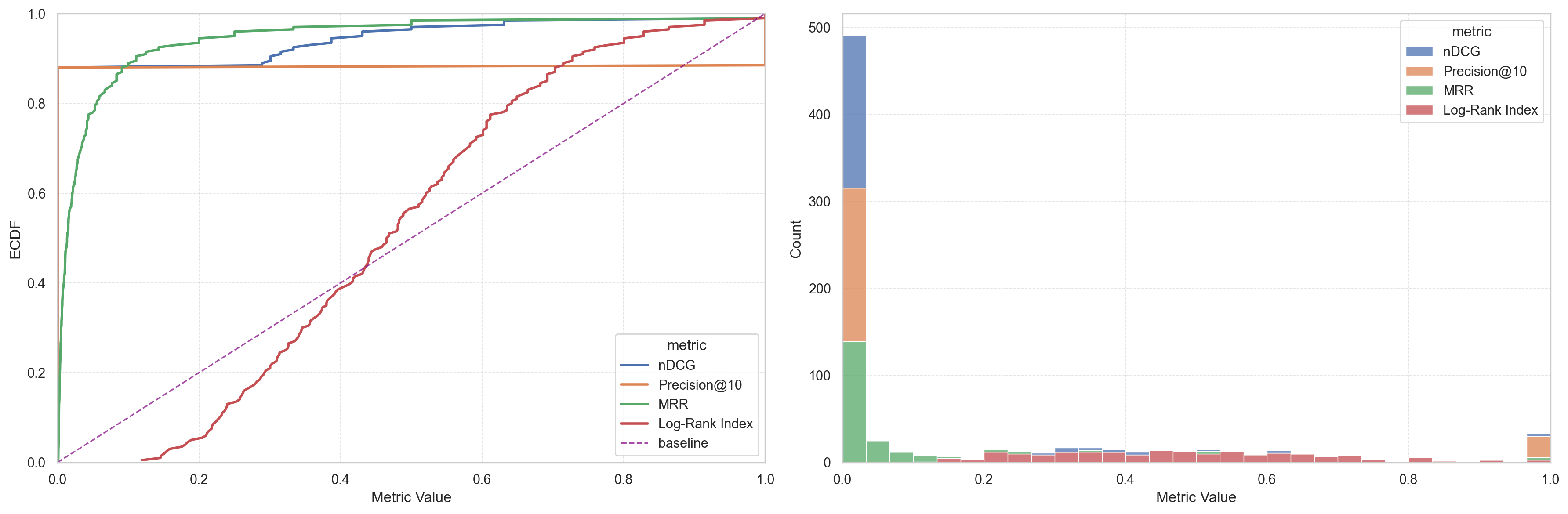}
  \caption{Left: an empirical cumulative distribution function (ECDF) for four metrics. Right: a histogram showing the distribution of metric values for each metric.}
  \label{fig:group_distribution}
\end{figure}

We take the 20-question subset of the MasQA: Texas Instruments dataset as the benchmark set $D=\{(q_i,c_i)\}_{i=1}^{20}$, and use \texttt{text-embedding-ada-002} for vector retrieval to obtain the retrieval rank $r_i$ of each query $q_i$. We first compare the vector retrieval rankings before and after HCA text augmentation . Then, we introduce 1–8 rounds of semantic perturbations (synonym replacements, noise insertion, etc.) to $q_i$ to construct eight degraded retrieval lists, yielding a total of 10 ranking sequences.
For each group, we compute four metrics: nDCG@10, Precision@10, MRR, and Log-Rank Index, and compare them from two perspectives: group mean trends (Figure~\ref{fig:group_mean}) and distribution shapes (Figure~\ref{fig:group_distribution}).

As the group index increases from 1 to 10 (with declining retrieval quality), nDCG@10 and Precision@10 drop sharply to 0 once relevant documents fall outside the top 10, leading to long plateaus and reduced discriminative power for medium- and low-quality retrievals. MRR decays as $1/r$, showing limited sensitivity in the mid-to-late range due to compression. In contrast, the Log-Rank Index remains smooth and monotonically decreasing across all groups, unaffected by the top-K cut-off, offering continuous feedback across all quality levels. As shown in Figure~\ref{fig:group_mean}, Log-Rank Index maintains a stable slope and strong group separability, reflecting its "first-order sensitivity" to rank changes.

From the query-level score distribution in Figure~\ref{fig:group_distribution}, nDCG@10 and Precision@10 show heavy accumulation near 0, with many zeros and a few values near 1, indicating discretization and degeneration. MRR is right-skewed, with low values as quality drops, resulting in poor differentiation in later groups. In contrast, the Log-Rank Index is evenly spread across $[0,1]$, with an ECDF closer to the diagonal and a stable histogram across intervals. This "non-cutoff, continuous, non-saturating" behavior allows Log-Rank Index to better reflect retrieval quality changes in RAG scenarios, providing more stable, differentiable feedback for downstream tuning and learning, such as weighted training or RL signals.

\section{Conclusion}


This paper introduces HiQA, a framework that tackles the challenge of distinguishing semantically and structurally similar documents in retrieval-augmented multi-document question answering. By enhancing text segments with hierarchical metadata, HiQA improves retrieval accuracy and answer quality in corpora where document titles, section paths, and domain entities provide useful disambiguating cues. Key components include a Markdown Formatter to preserve document structure, a Hierarchical Contextual Augmentor that cascades metadata into segments, and a Multi-Route Retriever combining vector-based, lexical, and keyword/entity matching.

Experiments on public datasets (NarrativeQA, Qasper) and the new MasQA benchmark show that HiQA is especially effective for structured and highly similar document collections, while remaining competitive rather than universally superior on public MDQA benchmarks. Our analyses show that cascading metadata “soft-partitions” the knowledge corpus, improving segment retrieval. The MasQA dataset, covering technical manuals, financial reports, and textbooks, highlights real-world QA challenges.

HiQA also integrates well with other retrieval-augmented methods, enhancing their performance. We hope that HiQA and MasQA inspire further development of metadata-centric RAG methods, enabling more reliable, adaptable systems for both academic and practical use.

\paragraph{Limitations and future work.}
HiQA depends on reasonably accurate structural extraction. If the Markdown Formatter misses headings, assigns wrong levels, or constructs incorrect parent-child paths, the augmented metadata can become noisy and may reduce retrieval precision. The method is also less helpful when documents have weak hierarchy, generic headings, or low entity density, where keyword/entity bonus may provide little benefit. Our table handling focuses on retrieval-oriented table metadata and still relies on the generator for numerical reasoning. Our image component is limited to image-reference retrieval and is not a full visual reasoning system. Finally, commercial-system comparisons are black-box and cannot be interpreted as controlled compute-equivalent baselines. Future work will focus on parser verification, adaptive score calibration, stronger multimodal document QA, and broader evaluation in low-structure domains.

\backmatter

\bmhead{Supplementary information}
The accompanying supplementary file \textit{
the appendices} include use-cases, auxiliary table and image processing details, detailed hyperparameter analysis, distributional exploration, and supplementary metric/model comparisons.




\section*{Declarations}


\begin{itemize}
\item Funding: This work is partially supported by the National Natural Science Foundation of China (6247072715).
\item Conflict of interest/Competing interests: The authors declare that they have no conflict of interest.
\item Data availability: \url{https://github.com/TebooNok/MasQA}
\item Code availability: \url{https://github.com/TebooNok/HiQA}
\item Author contribution: X.C. defined the research problem, implemented the solution, and contributed to the manuscript writing; P.G. selected the technology, conducted the literature review, and contributed to the writing; J.S. was responsible for the experimental implementation and figure preparation; X.C. provided the computational resources, built the dataset, and participated in the experiments; X.T. provided guidance on the methodology and manuscript writing. All authors reviewed the manuscript.
\end{itemize}









\begin{appendices}

\section{}
\begin{figure}[b]
    \centering
    \begin{subfigure}[b]{0.493\columnwidth}
        \includegraphics[width=\linewidth]{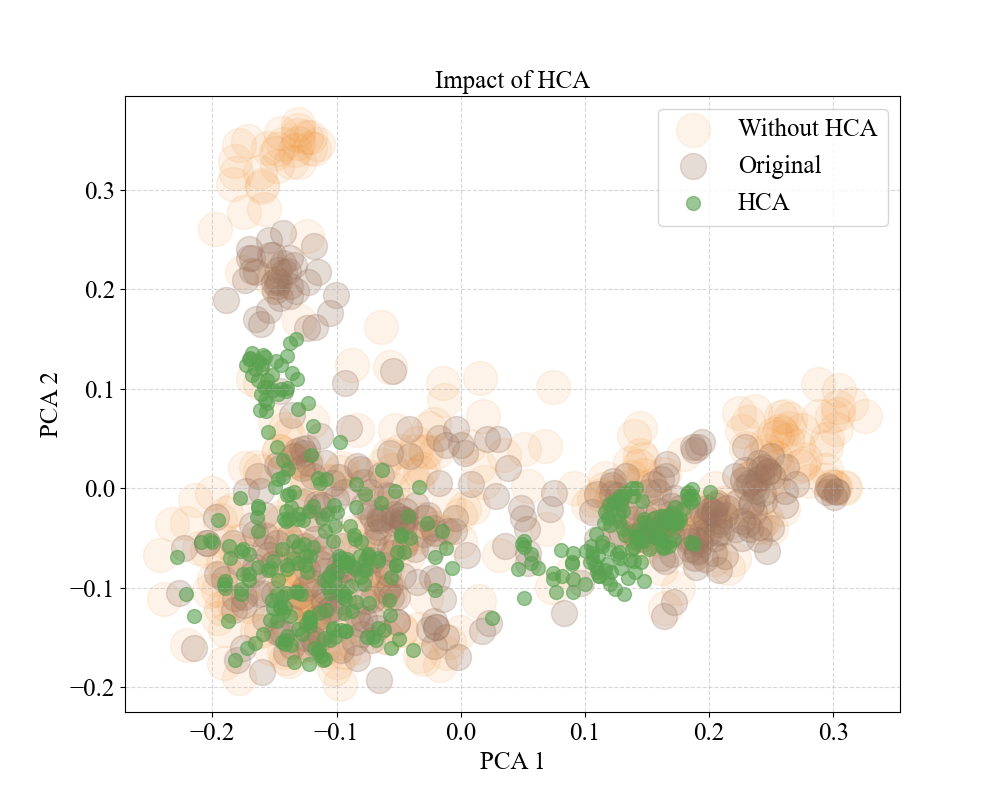}
        \caption*{(a)}
    \end{subfigure}
    \begin{subfigure}[b]{0.493\columnwidth}
        \includegraphics[width=\linewidth]{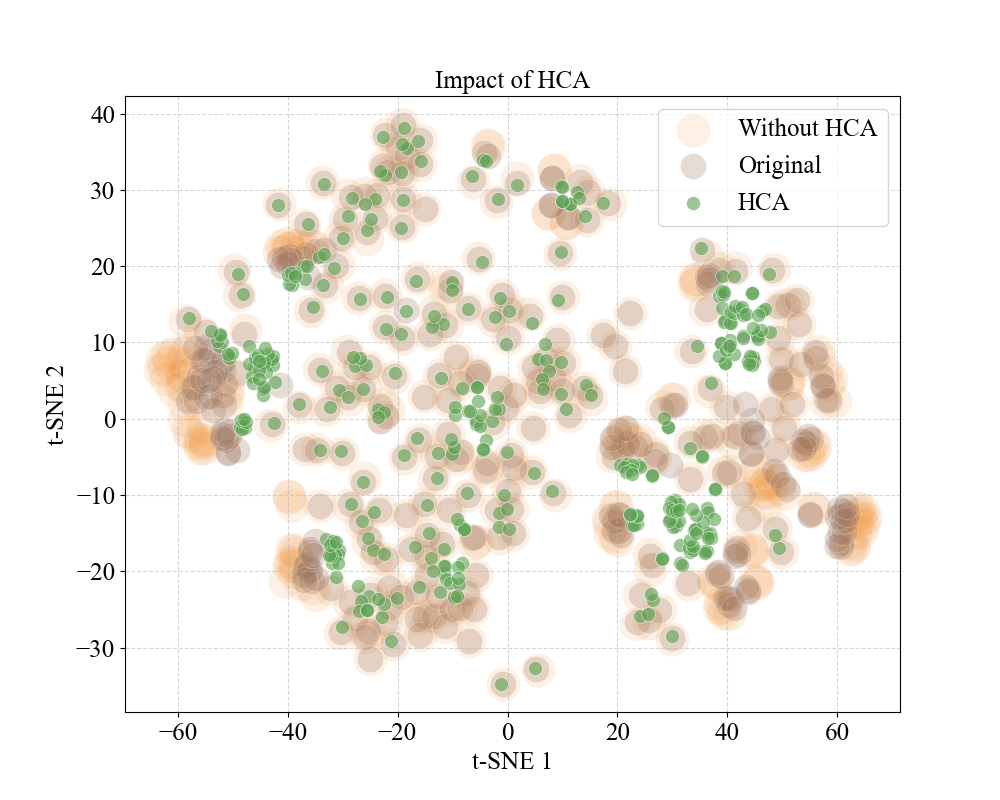}
        \caption*{(b)}
    \end{subfigure}
    \caption{Cohesion within Single Document.} (a) The figure illustrates the PCA visualization. (b) The figure depicts the t-SNE visualization. These two figures indicate that incorporating additional information into document segments results in a more compact spatial distribution of documents while maintaining stable intra-class distributions.
    \label{fig: single1}
\end{figure}

\begin{figure}[htbp]
    \centering
    \begin{subfigure}[b]{0.44\columnwidth}
        \includegraphics[width=\linewidth]{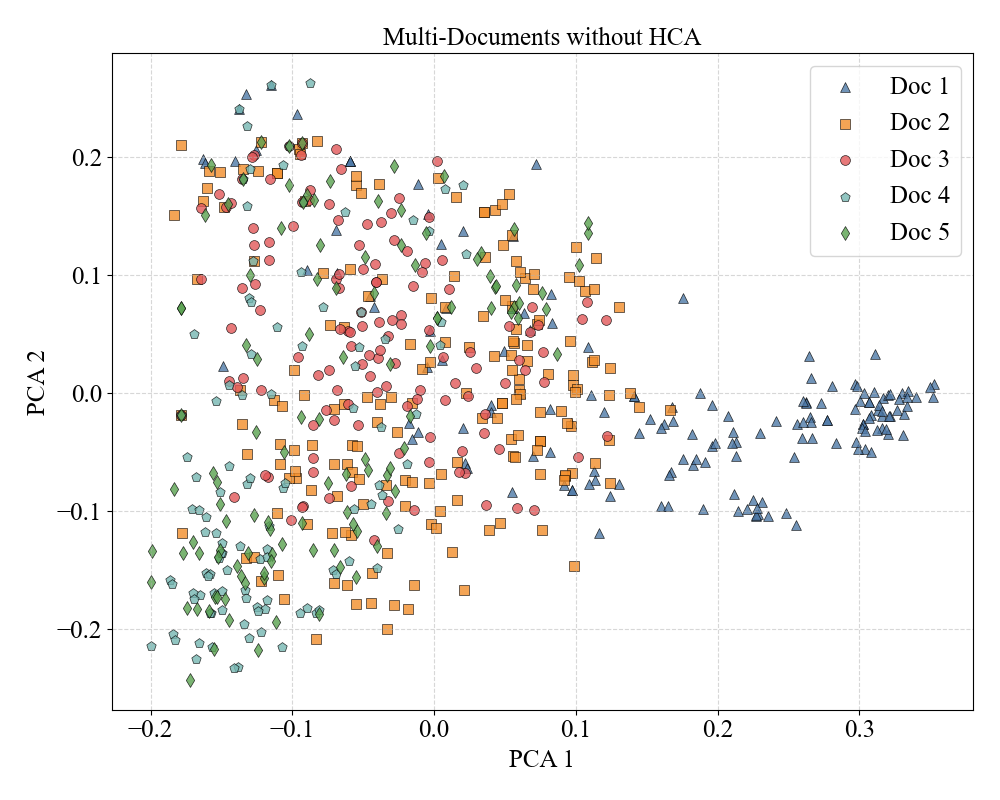}
        \caption*{(a)}
    \end{subfigure}
    \begin{subfigure}[b]{0.44\columnwidth}
        \includegraphics[width=\linewidth]{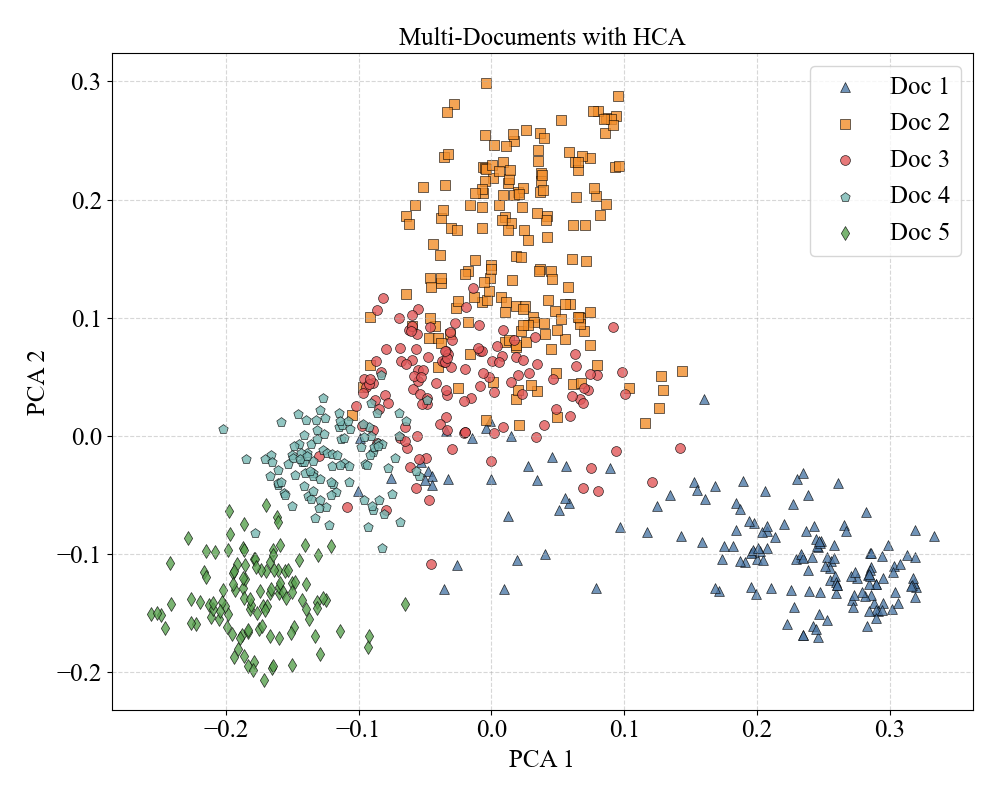}
        \caption*{(b)}
    \end{subfigure}
    \caption{Cohesion among Multi-Document. We visualize all points. (a) is distribution without HCA. After applying HCA (b) to incorporate hierarchical metadata into the segments of multiple documents, segments belonging to different documents exhibited strong cohesion and significantly reduced coupling, which clearly facilitates noise filtering.}
    \label{fig: multi}
\end{figure}

\begin{figure}[htbp]
    \centering
    \includegraphics[width=0.6\linewidth]{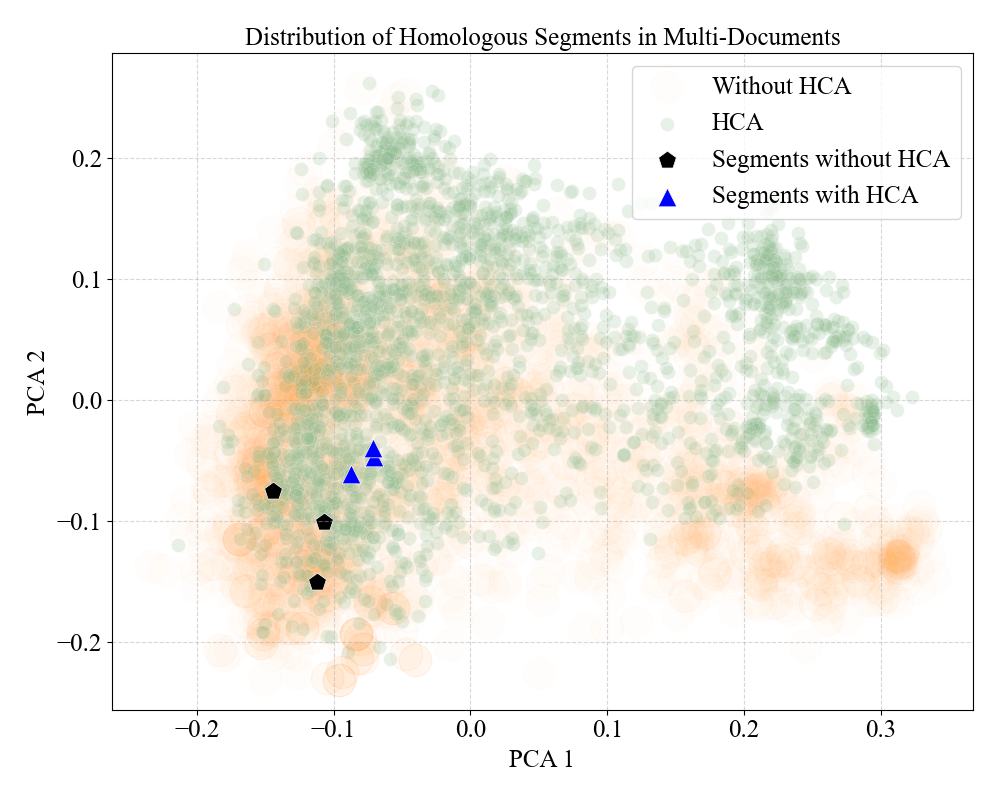}
    \caption{Cohesion among Homologous Sections. We visualized all segments from the documents and highlighted the "Application" sections of three selected documents. As shown, after HCA processing, content of the same type across different documents is more closely clustered, facilitating cross-document and multi-hop retrieval.}
    \label{fig: homo}
\end{figure}

\begin{figure}[htbp]
    \centering
    \begin{subfigure}[b]{0.44\columnwidth}
        \includegraphics[width=\linewidth]{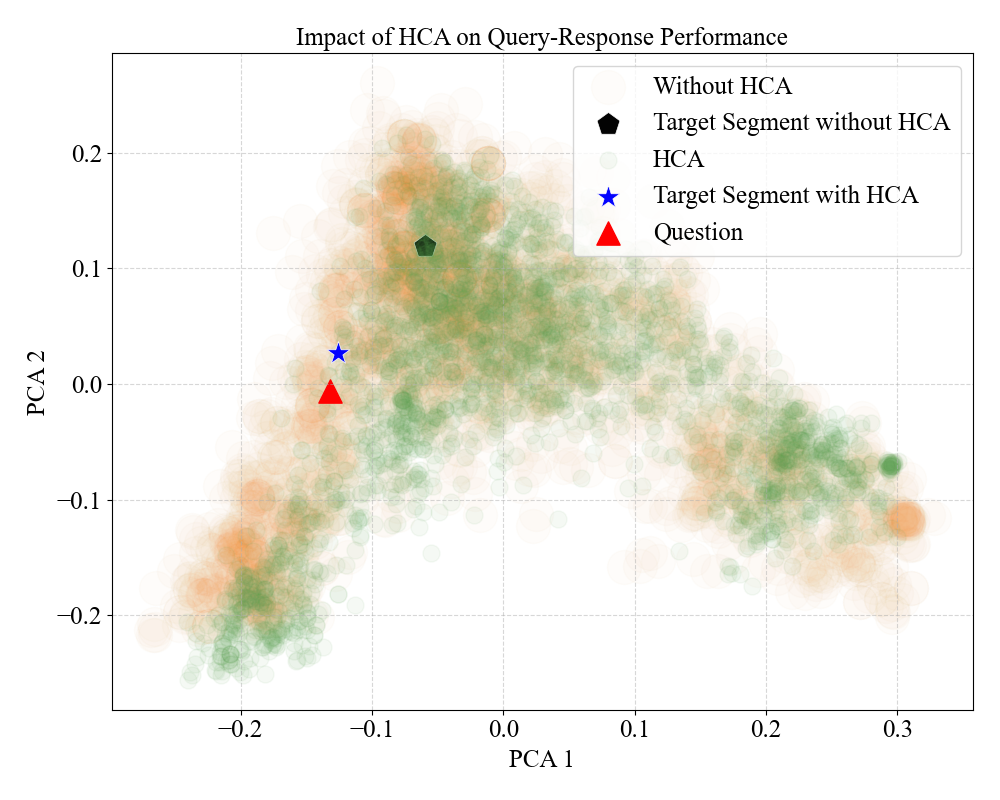}
        \caption*{(a)}
    \end{subfigure}
    \begin{subfigure}[b]{0.44\columnwidth}
        \includegraphics[width=\linewidth]{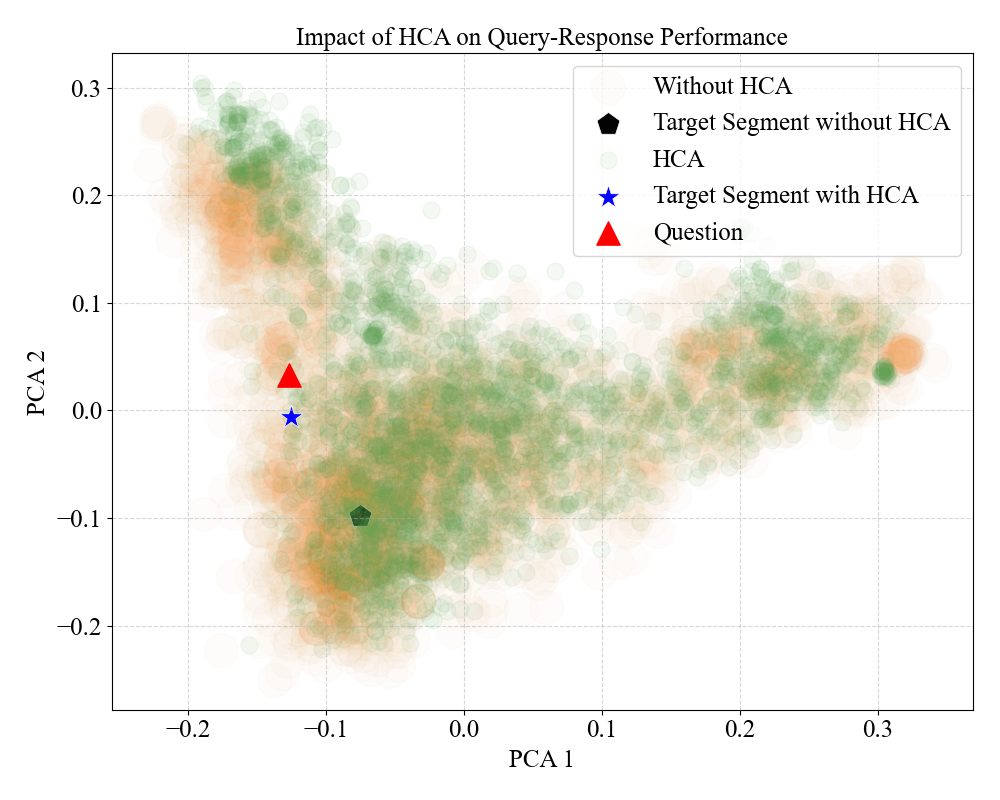}
        \caption*{(b)}
    \end{subfigure}
    \caption{Cohesion in Context Response. We selected two Question–Context pairs to evaluate the impact of HCA. In the figure, the black and blue points represent an identical knowledge segment before and after HCA; however, after HCA processing, the blue point is enriched with hierarchical metadata, which brings it closer to the question in the embedding space.}
    \label{fig: resp}
\end{figure}
\subsection{Distribution Exploration in Documents}
\label{app:distribution_explore}
In this section, we want to explore and demonstrate that HCA can reshape the distribution of document segments in the embedding space by strengthening the cohesion among segments and between questions and segments, raising a soft partition effect.

We quantitatively analyze distribution movements via PCA and tSNE visualization on a two-dimensional plane. The first three experiments focused on observing the impact of HCA on the distribution of document segments. The last experiment more specifically examined the spatial distribution of vector representations for given question-context pairs (Target Segment) in the embedding space.

\paragraph{RQ 1. How does HCA improve cohesion within a single document?}

We selected a document and applied three embedding processing methods: with HCA, i.e., enhancing segments with hierarchical metadata; with Original Segment, i.e., enhancing segments with the nearest title; and without HCA, which means no enhancement. Then, we compare the three sets of embedding vectors using PCA and tSNE. The results depicted in Figure \ref{fig: single1} (a) and (b) demonstrate that the implementation of HCA leads to a more compact distribution. These findings indicate that our approach can enhance the focus of the RAG algorithm on the target domain.

\paragraph{RQ 2. How does HCA improve cohesion among multi-document?}

We analyzed five documents from a dataset to compare their distributions with and without HCA. In a multi-document scenario, segments within each document naturally form a cluster. Thus, we can examine the distribution of these clusters. As illustrated in Figure \ref{fig: multi}, documents from the same dataset exhibit inherent similarities, leading to overlapping distributions and increasing retrieval complexity. However, data processed with HCA showed significant intra-cluster cohesion, effectively creating a soft partition of the documents which circumvents the information pruning associated with hard partitioning methods like Llamaindex.
\paragraph{RQ 3. How does HCA improve cohesion within homologous sections?}
We visualize all segment vectors from a dataset; then we highlight homologous sections across all documents in this dataset, e.g., all "Application" sections from each manual. As depicted in Figure \ref{fig: homo}, similar segments in different documents are observed to cluster more when processed with HCA, facilitating the answer of questions between documents. Specifically, the three blue dots represent the same chapter across three different documents. Although the chapters indicated by both the blue and black dots are identical, the content marked by the black dots does not include hierarchical metadata. When cross-document queries are made on this chapter, the distribution reveals that the relevant documents are more easily retrieved, which leads to higher recall under a fixed context size.
\paragraph{RQ 4. How does HCA improve cohesion in context response?}
We select a Question-Context pair. The question's embedding was marked on the visualization plane. Subsequently, the contexts processed with and without HCA were also plotted to observe their positions and distances relative to the question. As shown in Figure \ref{fig: resp}, our method significantly reduces the distance between the relevant context and question in the embedding space, greatly enhancing retrieval accuracy. This finding corroborates the substantial improvements observed in our method's Log-Rank Index.

\subsection{Proposed Question-Answering System}

\begin{figure*}[htbp]
    \centering
    \includegraphics[width=0.9\linewidth]{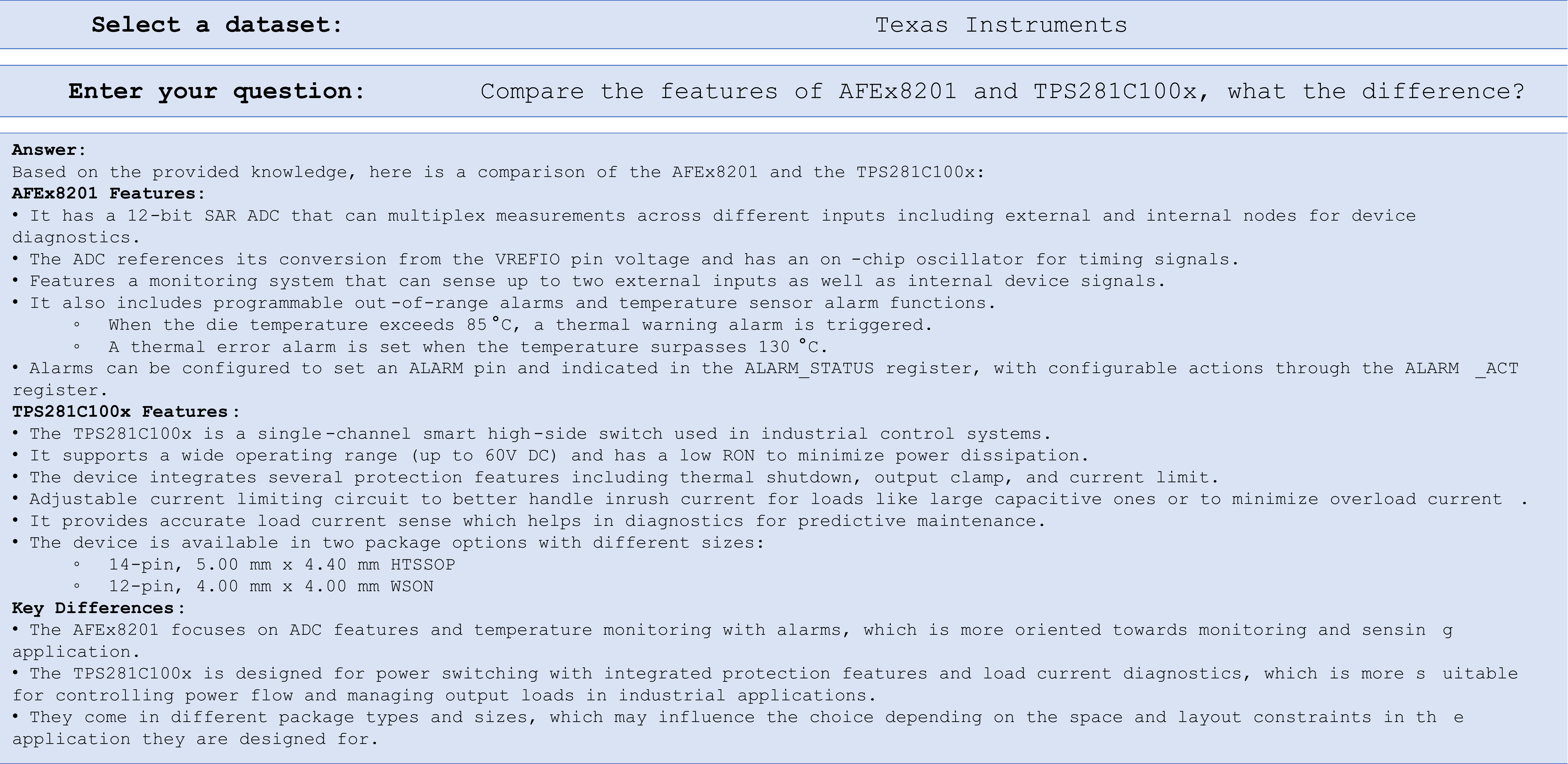}
    \caption{An Example Query-Answering on Texas Instruments Dataset}
    \label{fig:qa-example}
\end{figure*}

In the proposed framework, the question-answering process is single-stepped. Initially, relevant knowledge is retrieved from the document base using RAG according to the query. Subsequently, this context, in conjunction with the question, is fed into the language model to generate a response. The time taken to return the first character of the answer ranges between 1 to 3 seconds. An example of the QA process is illustrated in Figure \ref{fig:qa-example}.

\subsection{Image References in Responses}

\begin{figure*}[htpb]
    \centering
    \includegraphics[width=0.9\linewidth]{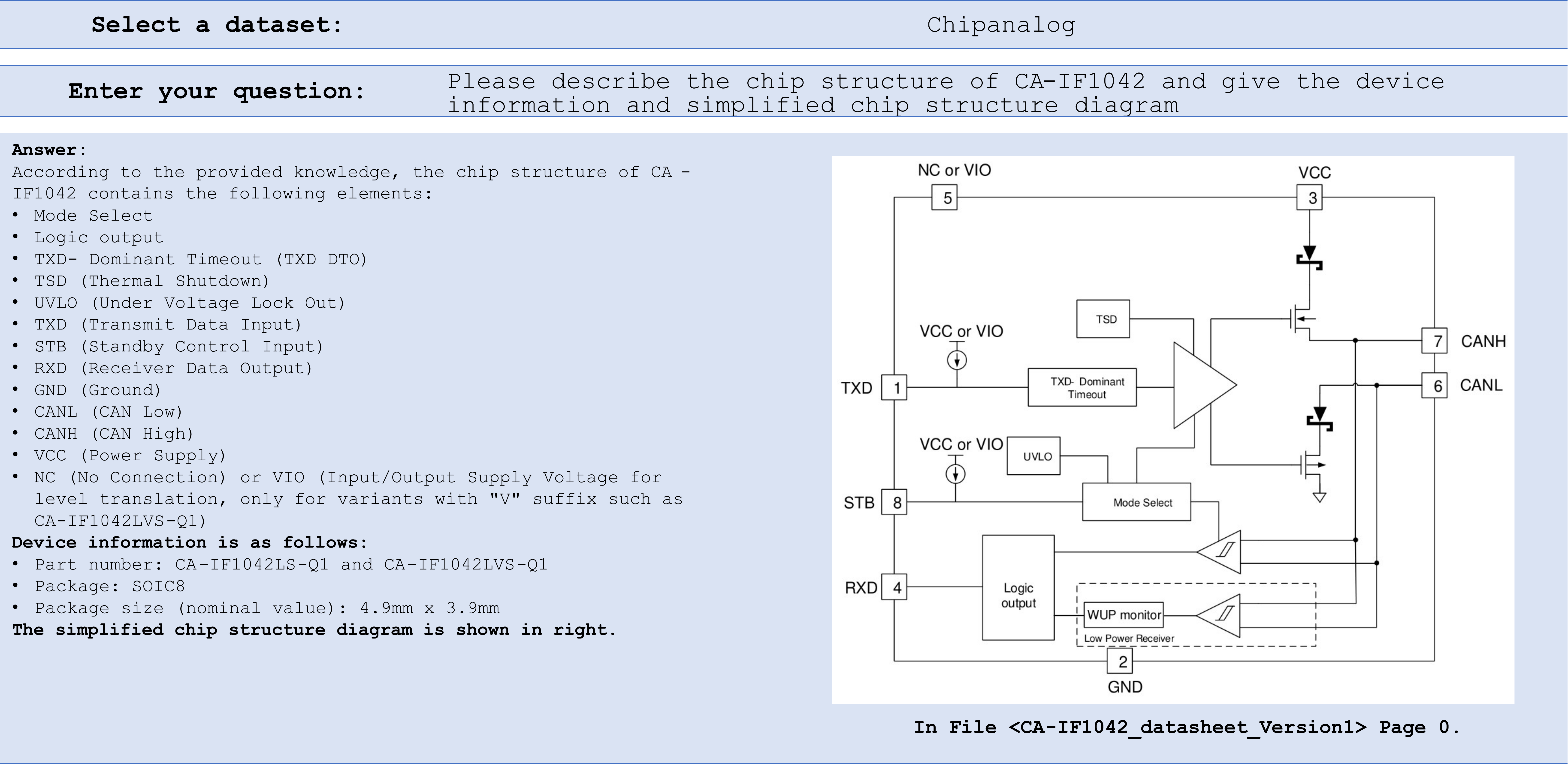}
    \caption{An Example Query-Answering via Image Reference}
    \label{fig:qa-image}
\end{figure*}

Our current system includes an auxiliary image-reference retrieval function, demonstrated in Figure \ref{fig:qa-image}. This extension retrieves image references using captions, surrounding text, and hierarchical context. It should be interpreted as a practical reference-retrieval feature rather than a fully evaluated multimodal visual reasoning module.

For images, we developed a tool named PDFImageSearcher, which is open sourced, to extract bitmap and SVG vector images from documents, as well as an API to retrieve an image. This utilizes the text surrounding the image, the image title, and an optional visual language model to generate a descriptive file for each image. Each document $D_M$ will have an image collection $D_G=\{I_1(\text{File}_1, \text{Desc}_1), I_2(\text{File}_2, \text{Desc}_2),...\}$.

\subsection{Table Augmentation}
\label{appendix:hadle_table}
\begin{figure}[htbp]
  \centering
  \includegraphics[width=0.8\linewidth]{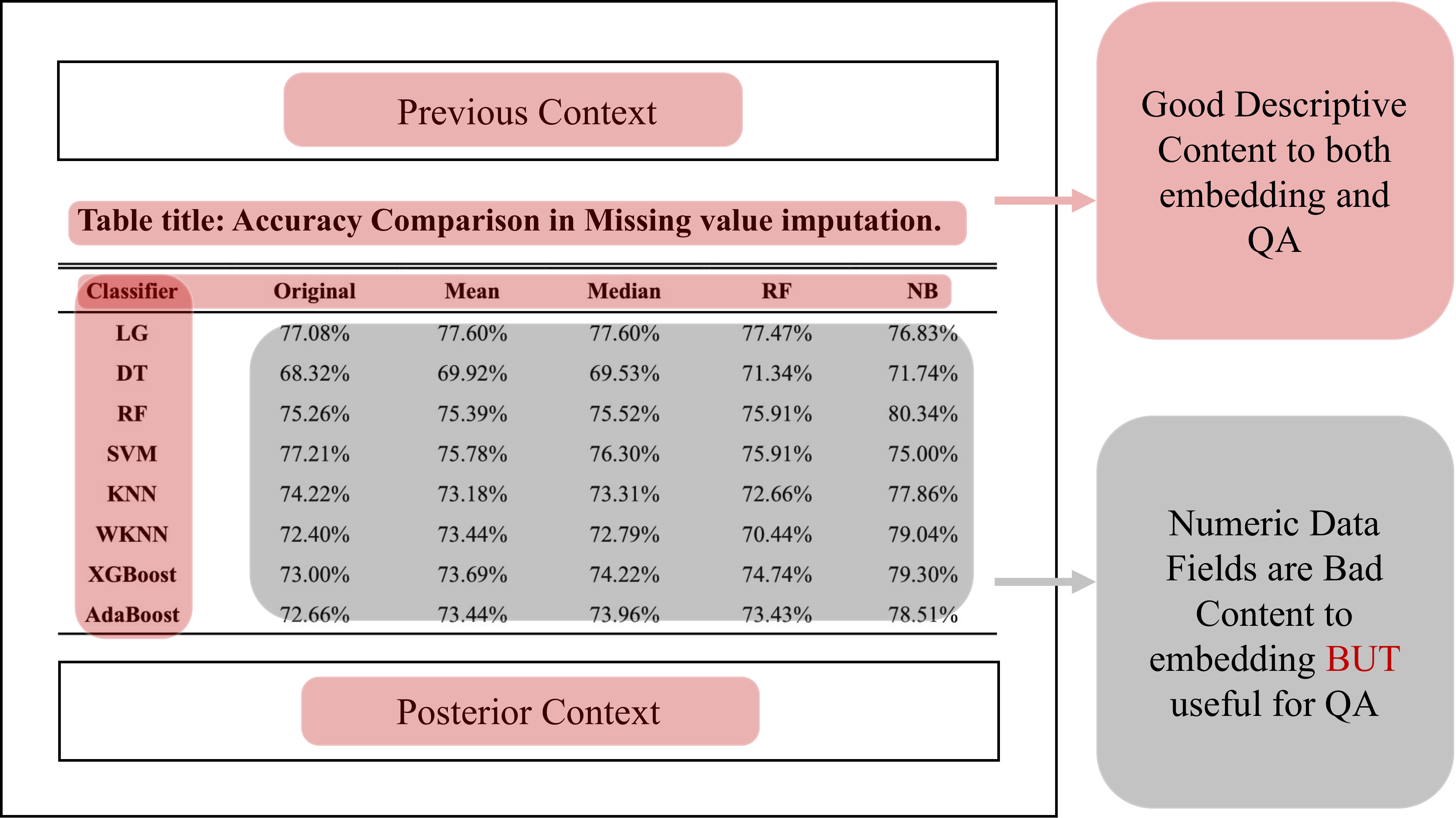}
  \caption{Embedding for Tables. Data fields are omitted to reduce noise during embedding. But if retrieved, these data fields are retained to provide context for LLMs}
  \label{fig:table}
\end{figure}

Traditional chunk-based RAG methods do not specifically address tables. Our experiments indicate difficulties in accurately recalling table information, largely because the numerical values in tables often behave as noise in semantic encoding. An example question is: "Does this phone have a 13,000 mah battery charge?". Actually, we need to match the battery rather than the number and use the retrieved number to fact-check. We posit that the semantic value of a table originates from its definition, including overall description, title, and row/column labels, as illustrated in Figure \ref{fig:table}. Hence, in embedding tables, we focus solely on these semantic elements, treating tables akin to text knowledge.

\subsection{Image Augmentation}
\begin{figure}[htbp]
  \centering
  \includegraphics[width=0.8\linewidth]{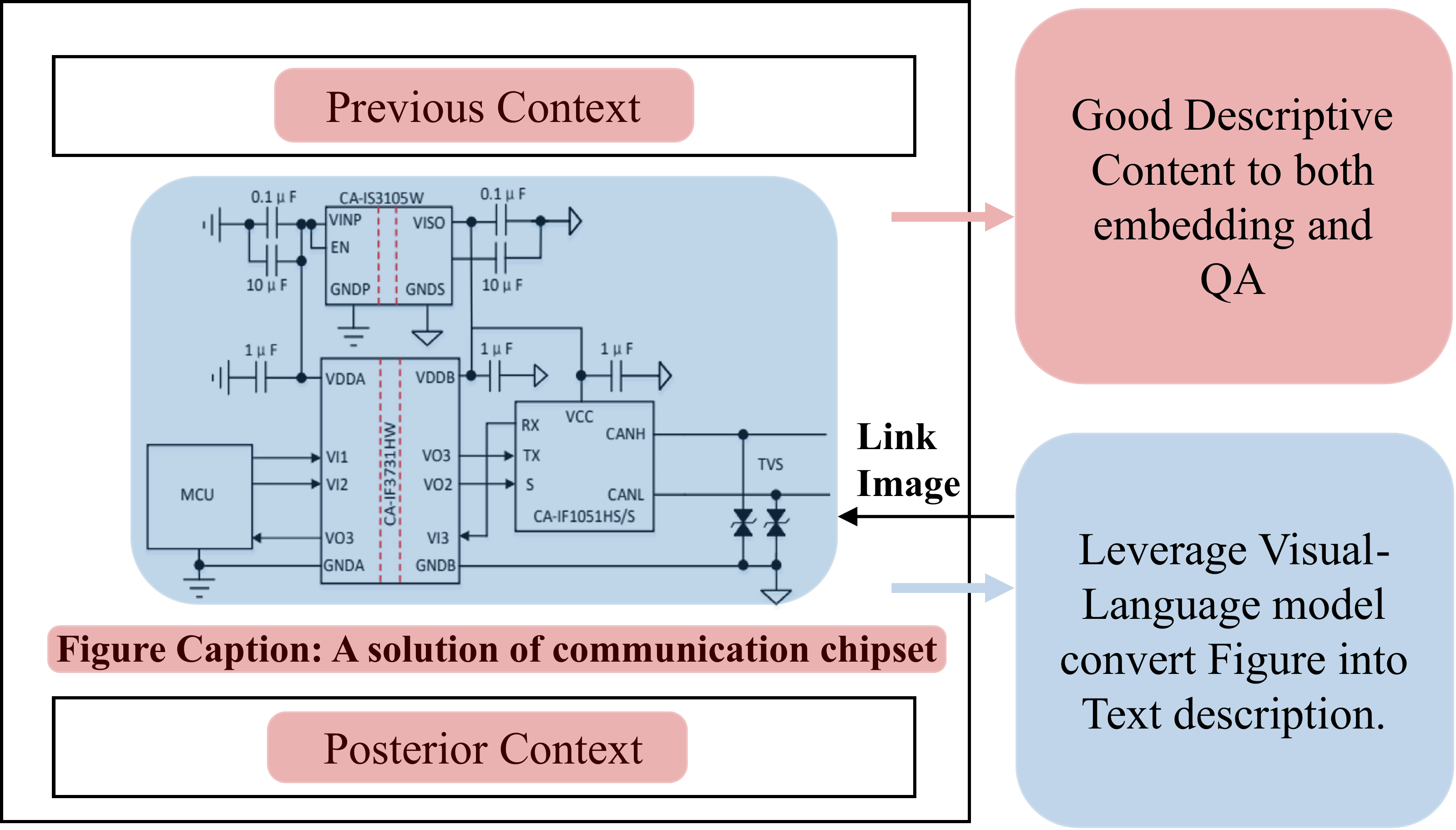}
  \caption{Embedding for Images. Applying a Visual-Language model to generate textual descriptions of the image semantics, which are then incorporated into the segment.}
  \label{fig:image}
\end{figure}

We utilize the wrapped context of an image and can further leverage visual language generation models to create descriptive captions that summarize visible content. These captions are embedded for image-reference retrieval. Because the present revision does not include a dedicated quantitative visual QA benchmark, we treat this module as an auxiliary extension and do not use it to support the main empirical claims. Image augmentation is shown in Figure \ref{fig:image}.

\subsection{Impact of Data Field Removal on Retrieval of Table}
\label{appendix:evaluate_removal}
\begin{figure}[htbp]
    \centering
    \includegraphics[width=0.8\linewidth]{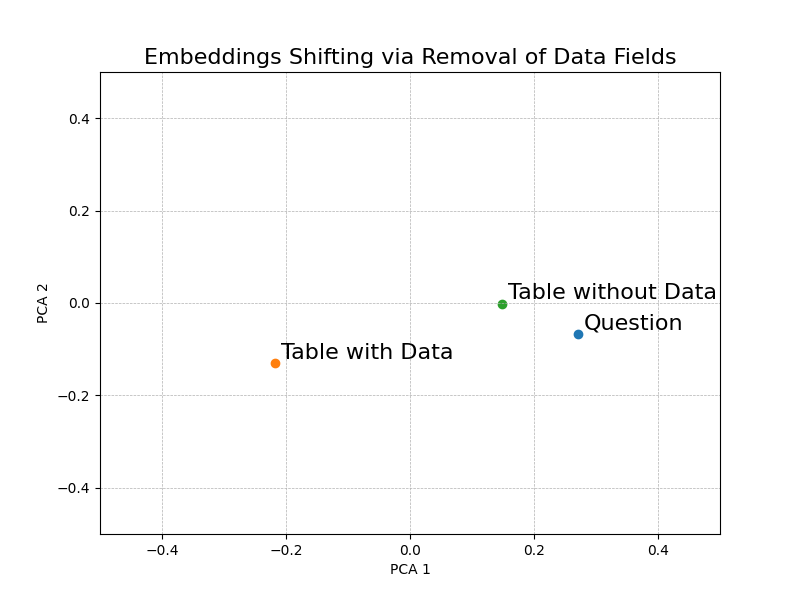}
    \caption{Embedding Shifting via Removing Data Fields of Table}
    \label{fig: removal}
\end{figure}

We examined the impact of removing data fields from tables during the embedding stage on the RAG method. As demonstrated in Table \ref{tab: removal} and Figure \ref{fig: removal}, embedding table captions and row/column labels while preserving the original numerical values for generation increases the inner product of context and question in the embedding space and reduces their distance in this space. This supports our table-oriented retrieval design, although complex table calculation remains dependent on the generator after the correct table is retrieved.

\begin{table}[htbp]
\centering
\begin{tabular}{@{}lcc@{}}
\toprule
Method & Table & Data Fields Removal \\
\midrule
Inner product & 0.879 & 0.913 \\
\bottomrule
\end{tabular}
\caption{Inner Product of Embedding between Question and Table Content}
\label{tab: removal}
\end{table}

\subsection{From IR to RAG: Why traditional metrics fall short}
\label{appendix:why_tradition_fall}
\begin{itemize}
\item \textbf{Scale of the document collection.} Typical information retrieval (IR) systems operate on collections of millions or billions of documents, which motivates the use of metrics restricted to the top-$K$ results. If a relevant document falls outside the top-$K$, its contribution to the score is truncated to zero. In contrast, the corpora used for retrieval in RAG are orders of magnitude smaller; a suitable metric should evaluate the whole ranked list without discarding documents.
\item \textbf{Strictness of relevance.} In IR, relevance is usually graded and metrics emphasise semantic similarity. RAG, especially in open-domain applications, requires exact matching of ground truth facts to prevent hallucinations in generation. Consequently, minor rank differences are more important than absolute recall.
\item \textbf{Context window constraints.} Large language models have bounded context windows. Only a limited number of retrieved chunks can be passed to the generator. A RAG metric should therefore favour retrieving relevant documents as early as possible and provide a continuous reward for lower ranks rather than a hard cut-off as in traditional top-$K$ metrics.
\end{itemize}


\subsubsection*{\textbf{Limitations of existing IR metrics from a single target chunk perspective}}

We consider the scoring function $S(r_i)$ for a single relevant chunk at rank $r_i$ returned by the retriever. Conventional IR metrics compute $S(r_i)$ differently:
\begin{itemize}
\item \emph{Precision@K:} \cite{schutze2008introduction} Precision@K counts the fraction of relevant documents among the top $K$ results and does not distinguish their order. It can be expressed for a single relevant document as $S_{\text{P@K}}(r_i)=\mathds{1}(r_i\leq K)$. Increasing $K$ to the corpus size $N$ makes this metric constant and insensitive to ranking; conversely, restricting to a small $K$ yields zero whenever $r_i>K$.
\item \emph{Mean reciprocal rank (MRR):} \cite{voorhees1999trec} MRR emphasises the first relevant document and disregards all subsequent relevant documents. For a single document it is $S_{\text{MRR}}(r_i)=1/r_i$. While this encourages high rankings, the reciprocal decreases rapidly and is insensitive to lower ranks; moreover, only the first relevant document contributes to the score.
\item \emph{Normalized discounted cumulative gain (nDCG):} \cite{jarvelin2002cumulated} DCG assigns each document a score $S_{\text{DCG}}(r_i)1/\log_2(r_i+1)$, which decreases logarithmically with rank, and nDCG normalizes this by the score of an ideal ranking at cut-off $K$. Although the logarithmic discount captures diminishing importance with rank, nDCG is computed relative to an ideal top-$K$ ranking and truncates contributions beyond $K$.
\end{itemize}

\subsection{Dataset Composition and Construction}
\label{appendix:dataset_compositon}

Existing datasets generally assume a clear distinction between relevant and non-relevant documents, overlooking the difficulty of retrieving precise information when multiple documents share substantial semantic overlap. While HotpotQA incorporates multi-hop reasoning across large-scale sources, it does not explicitly address retrieval difficulties arising from highly similar documents. Similarly, NarrativeQA~\cite{kocisky2018narrativeqa} and Qasper~\cite{dasigi2021qasper} introduce long-document reasoning tasks, but they do not evaluate retrieval robustness in environments where document distinction is ambiguous. This limitation is particularly problematic in large-scale retrieval scenarios, where models must discern subtle differences between sources to avoid generating misleading or redundant responses.

\begin{figure}[ht]
  \centering
  \includegraphics[width=0.8\linewidth]{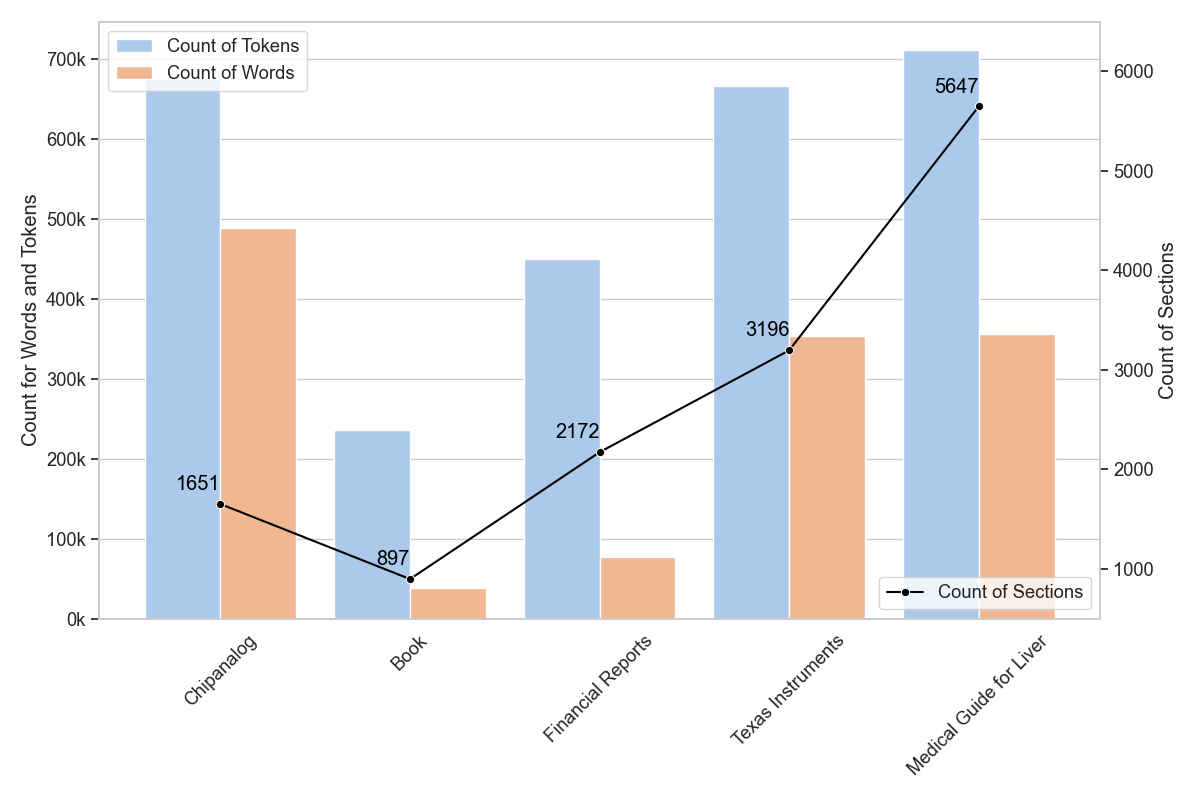}
  \caption{Statistical Information on the Scale of the Dataset: While typical RAG applications operate on datasets comprising fewer than 100 chunks, the MasQA dataset is substantially larger compared to other MDQA datasets, underscoring both the challenges and the practical implications.}
  \label{fig: dataset}
\end{figure}

To bridge this gap, MasQA primarily focuses on the challenges of distinguishing between documents with high similarity in multi-document RAG scenarios. Although the dataset is considered large compared to single-document RAG approaches, it may not compare in size to widely-used datasets such as HotpotQA or Wikipedia. However, these larger datasets do not face the same challenge of high similarity between documents, which is the intention of MasQA. Additionally, MasQA enables the evaluation of text enhancement techniques as a means to improve retrieval accuracy and question-answering performance. By providing a benchmark that explicitly incorporates indistinguishable multi-document challenges, MasQA facilitates the development of more robust retrieval mechanisms and enhances the applicability of RAG models in complex real-world scenarios. 

The MasQA dataset includes five distinct subsets, each specifically designed to represent different document scenarios. This variety ensures a thorough evaluation of RAG performance across diverse contexts and demonstrates its potential for real-world applications. Figure \ref{fig: dataset} provides a detailed comparison of these datasets across multiple
dimension.
The questions in the dataset were designed by expert users familiar with the respective domains. Most of the datasets, including Texas Instruments, Chipanalog, Analog Textbook, and Financial Report, were derived from a project aimed at developing an intelligent customer service system. Expert users in this project authored the questions, which were then used to evaluate the system's performance. The Medical Guide dataset, on the other hand, was created with questions generated by hospital doctors involved in another project. The question design reflects real-world use cases and common inquiry styles, ensuring that the RAG QA system is evaluated fairly and accurately.
\begin{itemize}
    \item \textbf{Technical Manuals from Texas Instruments} This subset includes 18 PDF files, each approximately 90 pages, featuring a mix of images, text, and tables in multiple languages.
    \item \textbf{Technical Manuals from Chipanalog} It consists of 88 PDF files, around 20 pages each, presented in a two-column format, enriched with images, text, and tables.
    \item \textbf{A College Textbook} A comprehensive 660-page book encompassing images, text, formulas, and tables.
    \item \textbf{Public Financial Reports Listed Companies} This consists of 8 reports for 2023, each report spans roughly 200 pages, mainly including text and tables.
    \item \textbf{Official Medical Guides for Liver} We collect 116 official liver diseases guides.
\end{itemize}

\paragraph{Question Bank}
For each subset, we crafted a question bank comprising question-answer-context triples. To show the application prospect of the proposed method, the questions are designed to mimic inquiries by engineers and analysts, covering various dimensions:
\begin{itemize}
    \item \textbf{Single and Multiple Choice Questions} Evaluating the dataset's capability to handle straightforward selection-based questions.
    \item \textbf{Descriptive Questions} Testing the ability to provide detailed explanations based on specific criteria.
    \item \textbf{Comparative Analysis} Involving multiple document segments for comparing several entities.
    \item \textbf{Table Questions} Assessing one or more tables extraction.
    \item \textbf{Across documents} Testing the ability to retrieve more than one document segment from multi-document.
    \item \textbf{Calculation} Testing the ability to gather information related to the questions and complete calculation problems.
\end{itemize}

Each question is annotated with correct answers and corresponding document segments. We will employ the Log-Rank Index for RAG metrics and assess the final answer quality to evaluate our methodology's efficacy in handling large-scale document bases and diverse document types.

\subsection{Question Bank Example}
\label{appendix:question_bank_example}
In ~\ref{tab:qa_example}, we show example questions for each question type.

\begin{table*}[htbp]
\begin{tabular}{|p{0.3\linewidth}|p{0.7\linewidth}|}
\hline

\textbf{Question Type} & \textbf{Examples} \\
\hline
Single Choice & Chipanalog 485 series interface products can support the highest rate is () A: 10Mbps B: 20Mbps C: 50Mbps D: 100Mbps
AFE7906's features of each DDC channels have A NCO (). A: 4 B: 8 C, 12 D: 16\\
\hline
Multiple Choice & Applications of ADC12QJ1600-SP are (). A: Electronic warfare (Signals intelligence, electronic intelligence) B: Satellite communications (SATCOM) C: Battery management systems D: Circuit breakers \\
\hline
Judgement & Statement: In the noise analysis of a common-gate amplifier, the input end of the circuit should be open when solving the equivalent input noise current source. True of False? \\
\hline
Descriptive & Please compare the production mode of the main business models of Zhejiang Huazheng and Weijie Chuangxin. \\
\hline
Comparison & Single choice: Which of the following product of Chipanalog is an ultra-low-power digital isolator? A: CA - IS3722HS B: CA - IS3742HW C: CS817x22HS D: CA - IS3841HW \\
\hline
Summary & What is the phase margin and what role does it play in system stability analysis? \\
\hline
Calculation & The parameters of the resistor-loaded common-source amplifier are: $ID = 100 \mu A$, $RD = 25 k\Omega$, MOS parameters are $VTHn = 1V$, $\mu nCOX =50 \mu A/V^2$, $\lambda=0.1V^{-1}$, $W/L=50/2$. [1.] Calculate the intrinsic transconductance $Gm$ (RS = 0$\Omega$, RL = 0$\Omega$) using the two-port model. \\
\hline
Crossing Documents & Fill the blank: Aiwei's non-current asset disposal profit and loss amount in 2022 is approximately () times that of Zhejiang Huazheng \\
\hline
Table Related & Single choice: The measurement range of OPT3004's characteristics is (). A: 0.001lux to 10lux B: 1lux to 15lux C: 0.01lux to 83,000lux D: 20lux to 83lux \\
\hline
\end{tabular}
\caption{Example of Question Bank}
\label{tab:qa_example}
\end{table*}

\subsection{Comparison of Different Models}
See table ~\ref{tab:qa_result}, we show and compare the generated output from different models.

\begin{table*}[htbp]
\centering
\resizebox{1\textwidth}{!}{
\begin{tabular}{>{\raggedright\arraybackslash}p{0.2\linewidth} 
                >{\raggedright\arraybackslash}p{0.15\linewidth} 
                >{\raggedright\arraybackslash}p{0.15\linewidth} 
                >{\raggedright\arraybackslash}p{0.15\linewidth} 
                >{\raggedright\arraybackslash}p{0.15\linewidth} 
                >{\raggedright\arraybackslash}p{0.15\linewidth}}
\toprule
\textbf{Question} & \textbf{Correct Answer} & \textbf{HiQA} & \textbf{GPT4} & \textbf{Llamaindex} & \textbf{ChatPDF} \\
\midrule
Single choice: In Q2 2022, which of the following companies had the highest net profit after deducting non-recurring gains? A: South Asia New Material B: Weijie Chuangxin C: Aiwei D: Guangdong Shengyi & D. & Based on the information provided, we can see that Guangdong Sheng Yi had the highest net profit after deducting non-recurring gains and losses in the second quarter. The answer is: D: Guangdong Shengyi. & In the second quarter, the net profit attributable to shareholders of listed companies after deducting non-recurring gains and losses was the largest. The answer is B: Weijie Chuangxin. & South Asia New Materials & The specific data of the four companies are not given. Therefore, I cannot answer this question accurately. \\
\addlinespace
Fill the blanks: The OPT3004’s Application Information includes two categories of interface: (), (). & electrical interface, optical interface & Electrical Interface, Optical Interface & Application information for the OPT3004 includes two interface categories: electrical and optical. & electrical and optical & The OPT3004’s Application Information includes two categories of interface: hardware and software. \\
\bottomrule
\end{tabular}}
\caption{Answers Across Different Methods}
\label{tab:qa_result}
\end{table*}

\subsection{Evaluate IR metrics with and without HCA}
\label{appendix:hca_metric}
\begin{figure}[htpb]
  \centering
  \includegraphics[width=0.8\linewidth]{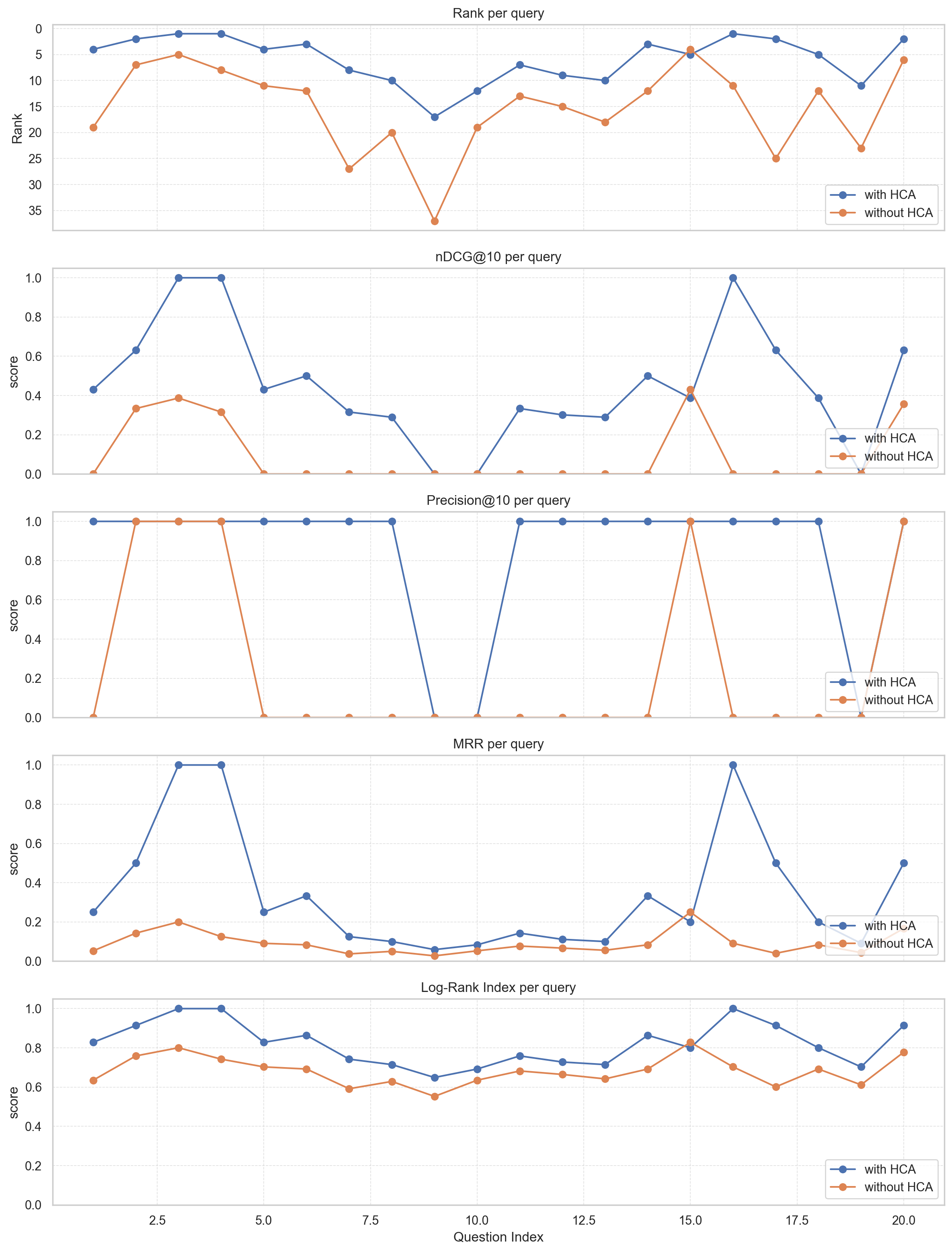}
  \caption{The figure presents a detailed comparison of the changes in rank for various metrics before and after the application of HCA. Each plot shows the performance of different metrics across different queries.}
  \label{fig:hca_rank_metric}
\end{figure}

As shown in Figure \ref{fig:hca_rank_metric}, on one hand, HCA significantly and consistently improves the position of the target chunk, suggesting that HCA contributes to enhancing the performance of the RAG system. On the other hand, traditional IR metrics fail to meet expectations in this scenario, as they struggle to effectively evaluate performance changes in the retrieval system. These IR metrics exhibit issues such as truncation and discontinuity. In contrast, the Log-Rank Index is relatively stable and continuous, demonstrating superior performance.

\subsection{Keyword and Entity Bonus Implementation}
\label{appendix:keyword_entity_bonus}
The keyword/entity bonus uses a compact document-level keyword set. We construct this set from three sources.

\begin{itemize}
    \item \textbf{LLM-based keyword extraction.} We use GPT-4o as the automatic extraction model. For each document, the document title is provided as input, and the model is instructed to return salient document-level keywords that can help identify the document in retrieval. This step is used to capture general entities and user-facing aliases.
    \item \textbf{Rule-based domain identifier extraction.} We apply simple corpus-specific rules to identify high-precision identifiers, including complete product models, disease names, company names, report names, and part numbers. These rules prioritize exact strings and preserve punctuation or alphanumeric patterns when they are part of the identifier.
    \item \textbf{Expert lexicon.} Domain knowledge is used to manually maintain one core discriminative keyword or identifier for each document. Examples include the main disease name in a medical guide, the complete product model in a technical manual, or the company name in a financial report.
\end{itemize}

The final keyword set is the union of these sources after normalization for whitespace and capitalization. During retrieval, a query and a candidate segment receive a keyword match count $m$ based on overlap with this set. The score contribution is normalized as $\log(1+m)/\log(1+M)$, where $M$ is the maximum possible or observed number of matched critical terms for the query. This keeps the bonus bounded and prevents it from dominating dense or lexical scores.

\subsection{Analysis and Guidence of Hyperparameters $\alpha$ and $\beta$}
\label{sec:hyper}
Regarding the Multi-Route Retriever and the tuning of hyperparameters $\alpha$ and $\beta$, here we provide a more comprehensive explanation of our experimental approach. These parameters are used to balance lexical and semantic aspects, as well as the level of professionalism of the user query.
\begin{figure}[htpb]
  \centering
  \includegraphics[width=0.95\linewidth]{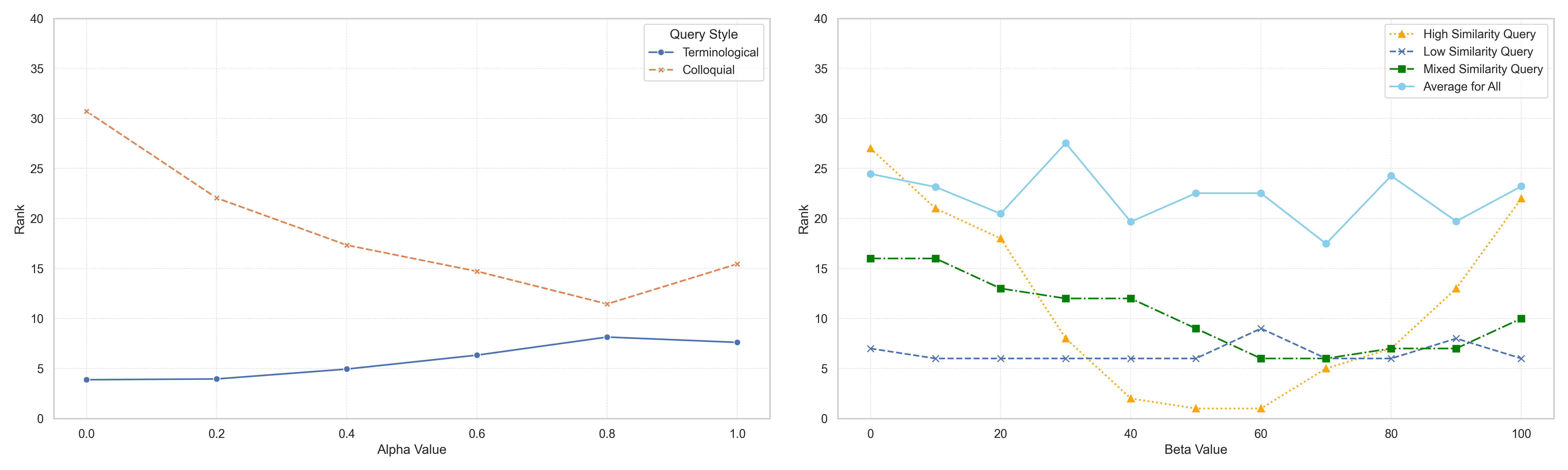}
  \caption{Left: Evaluation of varying $\alpha$ values under different query styles. Right: Evaluation of varying $\beta$ values under different similarity scenarios.}
  \label{fig:hyperparameter}
\end{figure}

\begin{figure}[htpb]
  \centering
  \includegraphics[width=0.95\linewidth]{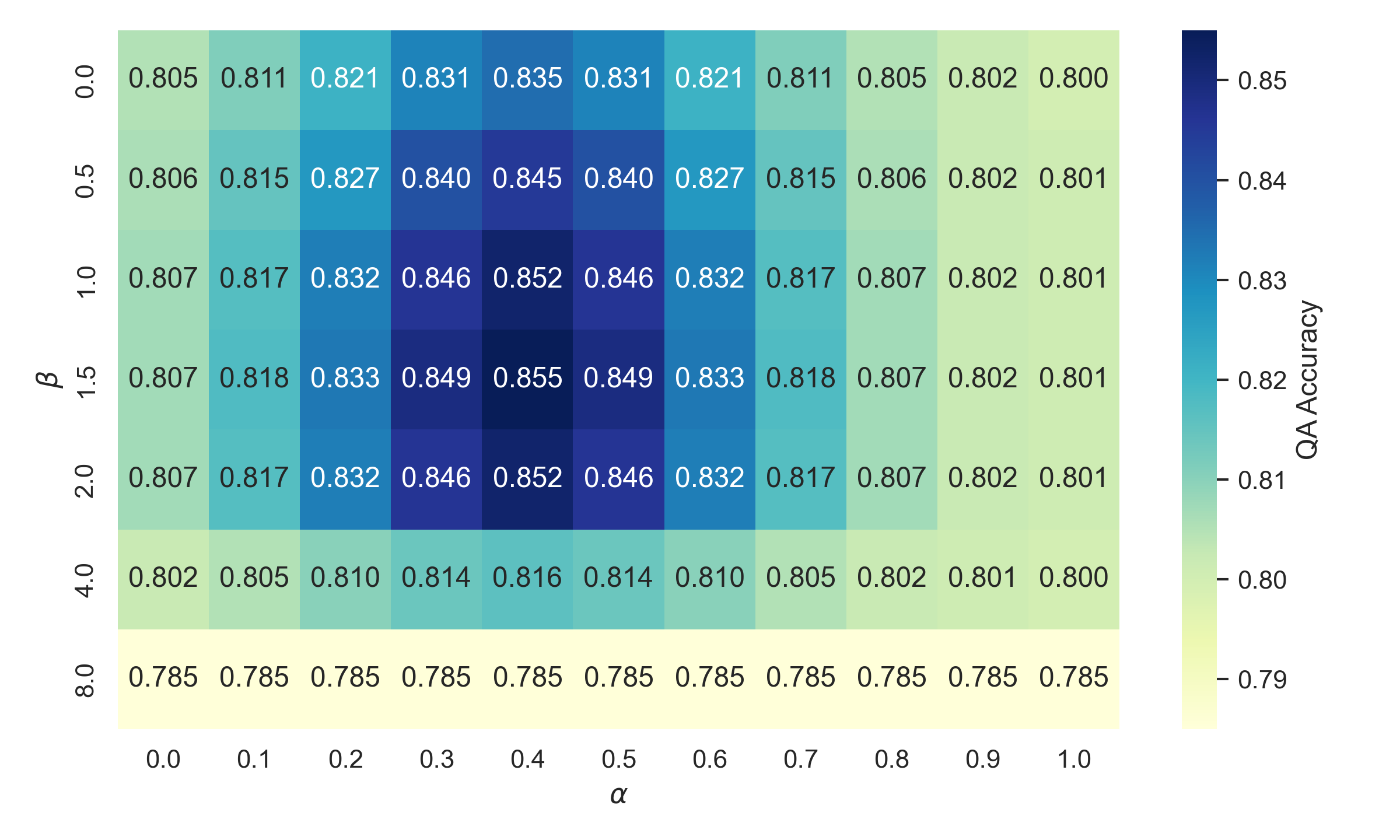}
  \caption{Sensitivity surface of normalized fusion. The best region appears when semantic and lexical scores are balanced and the normalized keyword/entity bonus is nonzero but not dominant.}
  \label{fig:alpha_beta_heatmap}
\end{figure}

We conducted experiments on the MasQA--Texas Instruments dataset, as illustrated in Figure~\ref{fig:hyperparameter}, the left figure presents the effect of varying $\alpha$ under two distinct query styles: one using precise, formal terminology and the other adopting more conversational phrasing. The results show that, for the terminological set, the average rank of the target document gradually deteriorates as $\alpha$ increases, reflecting the reliability of lexical matching when users employ domain-specific, accurate terms; in this case, smaller $\alpha$ values yield optimal performance. Conversely, for the colloquial set, the average rank improves markedly with increasing $\alpha$, indicating that semantic matching becomes more important when the query expression is less aligned with the document’s terminology. For example, in the query “\textit{Where is AMC131M02 typically used?}”, increasing $\alpha$ can raise the rank from 43 to 1 by better capturing the semantic association to the “Application” section, whereas for a clearly defined technical question such as “\textit{What are the two timing components of the turn-off transition in Turn-Off Times of Switching Parameters for the LMG342xR050?}” a larger $\alpha$ leads to a better rank. These patterns also imply that in cross-lingual scenarios, semantic-oriented retrieval preferences may emerge due to lexical mismatches.

The right panel of Figure~\ref{fig:hyperparameter} shows experiments that vary $\beta$, which controls the weight of the keyword bonus term $\beta \log(1+|C|)$ to emphasize critical keywords, particularly in corpora with high similarity of documents. While the overall average rank across all queries shows only minor fluctuations with changing $\beta$, substantial differences emerge in individual cases. We select three representative queries: one associated with many highly similar documents, one with few similar documents, and one with moderate similarity. The results demonstrate that keyword boosting can substantially improve retrieval in \textit{indistinguishable} multi-document settings; when the corpus contains numerous highly similar documents, increasing $\beta$ (e.g., in the range 20--40) helps surface the most relevant document by reinforcing the keyword–document connection. In contrast, in diverse corpora with low inter-document similarity, an excessively large $\beta$ can overemphasize keywords, potentially hindering standard retrieval channels.

\paragraph{\textbf{Practical guidance}} Based on these findings, we recommend dynamically adapting $\alpha$ and $\beta$ according to the formulation of the query and the characteristics of the corpus. For precise and terminology-rich queries, a small $\alpha$ (e.g. $<0.2$) with negligible keyword boost ($\beta \approx 0$) is preferable. For colloquial queries or corpora with high inter-document similarity, larger $\alpha$ and $\beta$ values (e.g., $\alpha>0.6$, $\beta\approx 60$) can enhance semantic matching and keyword discrimination. This adaptive tuning strategy enables the multi-route retriever to achieve robust performance across diverse retrieval scenarios.

\paragraph{Adjusting hyperparameters $\alpha$ and $\beta$}
\label{appendix:adjust_hyper}
For $\alpha$, we emphasize its flexibility and its dependence on the user's input style and the alignment with the knowledge base. For instance, when professional users use precise and formal terminology, a smaller $\alpha$ value is preferred, emphasizing lexical matching. In contrast, for casual users using more colloquial language, a larger $\alpha$ is more suitable, focusing on semantic matching. In our experiments, we demonstrate the effect of varying $\alpha$ through two sets of query modifications: one with more formal phrasing and another with more conversational language. As expected, using a larger $\alpha$ improves performance for the latter case.

For $\beta$, the tuning depends on the similarity between documents in the knowledge base. In scenarios where the document set is highly similar and large in quantity, increasing the keyword bonus (i.e., a higher $\beta$) helps distinguish relevant documents from nearly identical irrelevant ones. We add experimental evidence by comparing document sets with high and low similarity, where we show that a higher $\beta$ significantly improves the ranking in the highly similar set but has minimal impact on the less similar set.

\subsection{Comparative Analysis of LLM types Input Length}
In Table \ref{tab:model_accuracy}, we compare the accuracies of three models, GPT4, Moonshot, and ChatGLM-Pro, across different context lengths. This comparison aids in assessing the models' performance variations with token length changes. The results show significant differences between various models in the processing of noisy or long context information.

\begin{table}[htbp]
\centering
\begin{tabular}{@{}lcc@{}}
\toprule
\textbf{Model} & \textbf{Token Length} & \textbf{Accuracy (\%)} \\
\midrule
GPT-4o & 2k & 44/50 (88\%) \\
GPT-4o & 8k & 48/50 (96\%) \\
Moonshot & 2k & 42/50 (84\%) \\
Moonshot & 8k & 45/50 (90\%) \\
ChatGLM-Pro & 4k & 39/50 (78\%) \\
\bottomrule
\end{tabular}
\caption{Accuracy of Different Models with Various Token Lengths}
\label{tab:model_accuracy}
\end{table}

\end{appendices}

\end{document}